\documentclass[sigconf]{acmart}
\renewcommand\footnotetextcopyrightpermission[1]{} 

\usepackage[utf8]{inputenc} 
\usepackage[T1]{fontenc}    
\usepackage{url}            
\usepackage{booktabs}       
\usepackage{amsfonts}       
\usepackage{nicefrac}       
\usepackage{microtype}      
\usepackage{multicol}
\usepackage{color}
\usepackage{bm}
\usepackage{cuted}

\usepackage{mathtools}
\usepackage{multirow}
\usepackage{tikz}
\usetikzlibrary{matrix}
\usepackage{subfigure}
\usepackage{enumerate}
\usepackage{graphicx}
\usepackage{caption}
\usepackage[ruled,linesnumbered]{algorithm2e}
\usepackage{float}				  
\usepackage{enumitem}
\newtheorem{definition}{Definition} 
\newtheorem{theorem}{Theorem} 
\newtheorem{problem}{Problem}
\newtheorem{prop}{Proposition} 
\newtheorem{remark}{Remark}[section]

\usepackage{array}
\newcolumntype{L}[1]{>{\raggedright\let\newline\\\arraybackslash\hspace{0pt}}m{#1}}
\newcolumntype{C}[1]{>{\centering\let\newline  \\\arraybackslash\hspace{0pt}}m{#1}}
\newcolumntype{R}[1]{>{\raggedleft\let\newline \\\arraybackslash\hspace{0pt}}m{#1}}

\usepackage{pdftexcmds}
\usepackage{catchfile}
\usepackage{ifluatex}
\usepackage{ifplatform}
\usepackage{balance}
\usepackage{afterpage}
\usepackage{bbding}
\ifwindows
\usepackage{times}
\fi
\iflinux
\fi
\ifmacosx
\fi 

\AtBeginDocument{%
  \providecommand\BibTeX{{%
    \normalfont B\kern-0.5em{\scshape i\kern-0.25em b}\kern-0.8em\TeX}}}

\begin{document}

\title{Role-Aware Modeling for N-ary Relational Knowledge Bases}


\author{Yu Liu$^{12}$, Quanming Yao$^{23}$, Yong Li$^{12*}$}
\affiliation{%
 \institution{\textsuperscript{1}Beijing National Research Center for Information Science and Technology (BNRist)\\ \textsuperscript{2}Department of Electronic
 Engineering, Tsinghua University, Beijing 100084, China\\ \textsuperscript{3}4Paradigm Inc, Hong Kong}
}
\affiliation{$^*$Corresponding author: liyong07@tsinghua.edu.cn}
\renewcommand{\authors}{Yu Liu, Quanming Yao, Yong Li}
\renewcommand{\shortauthors}{Yu Liu, Quanming Yao, Yong Li}

\begin{abstract}
	N-ary relational knowledge bases (KBs) represent knowledge with binary and beyond-binary relational facts. 
	Especially, in an n-ary relational fact, 
	the involved entities play different roles, 
	e.g., the ternary relation \texttt{PlayCharacterIn} consists of three roles, \textsc{Actor}, \textsc{Character} and \textsc{Movie}. 
	However, 
	existing approaches are often directly extended from binary relational KBs, i.e., knowledge graphs, 
	while missing the important semantic property of \emph{role}. 
	Therefore, 
	we start from the role level, 
	and propose a Role-Aware Modeling, RAM for short, 
	for facts in n-ary relational KBs. 
	RAM explores a latent space that contains basis vectors, 
	and represents roles by linear combinations of these vectors. 
	This way encourages semantically related roles to have close representations. 
	RAM further introduces a pattern matrix that captures the compatibility between the role and all involved entities. 
	To this end, it presents a multilinear scoring function to measure the plausibility of a fact composed by certain roles and entities.  
	We show that RAM achieves both theoretical full expressiveness and computation efficiency, 
	which also provides an elegant generalization for approaches in binary relational KBs.	
	Experiments demonstrate that RAM outperforms representative baselines on both n-ary and binary relational datasets.
\end{abstract}




\maketitle
\section{Introduction} \label{sec:intro}
Traditional binary relational knowledge bases (KBs, a.k.a. knowledge graphs) describe real-world knowledge in a triple form, $(r,h,t)$, 
where $r$ is binary relation, $h$ and $t$ are head and tail entities respectively. 
However, 
recent studies find that considerable knowledge is beyond triple representation, 
involving with beyond-binary relations and more than two entities \citep{nickel2015review,hogan2020knowledge,rouces2015framebase}. 
These findings raise the n-ary relational KB modeling problem \citep{wen2016representation,guan2019link,liu2020generalizing,guan2020neuinfer,rossobeyond,fatemi2019knowledge}, 
representing binary and beyond-binary relational facts together, 
which is more general for human-level intelligence \citep{hogan2020knowledge,nickel2015review,ji2020survey}.

\begin{figure}[t]
	\centering
	\includegraphics[width=0.99\linewidth]{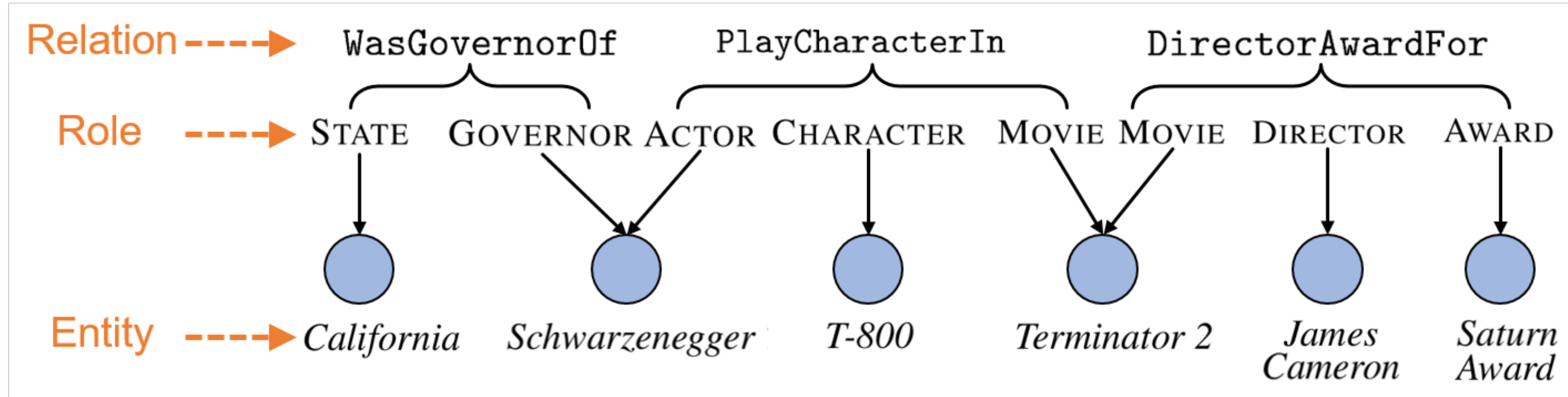}
	\vspace{-10px}
	\caption{Exampled n-ary relational facts in role-entity form, from Wikipedia\protect\footnotemark. 
		Each relation consists of semantic roles, which can be explicit (as in this figure) or implicit.}
	\label{fig:example}
\end{figure}

Compared with the simple binary relation, 
the n-ary relation describes the relationship between $n\geq2$ entities, 
containing much more complicated semantics. 
To address this, 
World Wide Web Consortium (W3C\footnote{\url{https://www.w3.org}}) and Schema.org\footnote{\url{https://schema.org/}} introduce the semantic property of \emph{role}\footnote{Note that the role here is conceptually different from the role concepts in other research domains, which are discussed in Appendix~\ref{app:role_work}.} for n-ary relations \citep{roleSchema}. 
For example, in Figure~\ref{fig:example}, a fact with ternary relation \texttt{PlayCharacterIn} is represented in role-entity form, 
$\{\textsc{Actor}\!\!:\!\textit{Schwarzenegger},\ \textsc{Character}\!\!:\!\textit{T-800},\ 
 \textsc{Movie}\!\!:\!\textit{Terminator 2}\}$. 
The roles identify semantics of involved entities to the n-ary relation, 
i.e., the \emph{star} and the \emph{political} semantics of \textit{Schwarzenegger} by roles \textsc{Actor} and \textsc{Governor}. 
Thus, the roles are crucial for n-ary relational KB modeling.

\footnotetext{\url{https://en.wikipedia.org/wiki/Arnold_Schwarzenegger}}

However, several existing approaches \citep{fatemi2019knowledge,wen2016representation,zhang2018scalable,liu2020generalizing} largely miss the roles, 
and still follow the relation level modeling in binary relational KBs. 
Especially, they embed the n-ary relation and entities into low-dimensional space, 
and measure the fact plausibility based on such embeddings. 
For example, both m-TransH \citep{wen2016representation} and RAE \citep{zhang2018scalable} extend binary relational model TransH \citep{TransH}, and project entities onto relation-specific hyperplanes for plausibility score, 
while HypE \citep{fatemi2019knowledge} extends SimplE \citep{kazemi2018simple} with multilinear product and positional filters, 
and GETD \citep{liu2020generalizing} extends TuckER \citep{balazevic2019tucker} in binary relational KBs. 
These models completely lose role semantics \citep{StarE,guan2020neuinfer}, 
and the first two are theoretically weak expressive \citep{kazemi2018simple,liu2020generalizing}. 
On the other hand, 
the role-entity form is adopted in recent works of NaLP \citep{guan2019link}, HINGE \citep{rossobeyond} and NeuInfer \citep{guan2020neuinfer}, 
which all leverage neural networks to measure the plausibility. 
Nonetheless, these models learn each role independently, 
ignoring the semantic relatedness among roles, 
e.g. \textsc{Actor} and \textsc{Movie}. 
The compatibility between the role and all involved entities are also missed, 
e.g., \textsc{Movie} with \textit{Terminator 2}, \textit{Schwarzenegger} and \textit{T-800}. 
Moreover, enormous parameters in neural networks also lead to high complexity and overfitting \citep{liu2020generalizing,kazemi2018simple}. 
Table~\ref{tab:MultiAry_SF} presents a comparison of existing approaches.

According to existing works on both binary and n-ary relational KBs, 
some valuable points on KB modeling can be obtained,  
i.e. a good KB modeling approach should: 
(1) be role-aware for complicated semantics \citep{guan2019link,guan2020neuinfer,rossobeyond}, 
including the semantic relatedness among roles and role-entity compatibility;  
(2) be fully expressive to represent all types of relations, 
e.g., the symmetric relation \texttt{Spouse} and the inverse relations of \texttt{ParentOf} and \texttt{ChildrenOf} \citep{kazemi2018simple,liu2020generalizing,fatemi2019knowledge,balazevic2019tucker,rossi2020knowledge}, 
and (3) be linear in both time and space complexity in order to scale to the growing size of current KBs \citep{bordes2013translating,kazemi2018simple,lacroix2018canonical,nickel2015review,trouillon2016complex}. 
To the best of our knowledge, 
none of existing approaches satisfy above three points, 
and it is still an open issue to be addressed.


Instead, in this paper, 
we focus on the unique characteristics in n-ary relational fact, 
i.e., the role, 
and propose \underline{R}ole-\underline{A}ware \underline{M}odeling, RAM for short, for n-ary relational KBs. 
Different from previous approaches
which are mainly extended from binary relational KB ones, 
RAM introduces a latent space for roles, 
where semantically related roles are supposed to share close representations. 
Moreover, RAM learns a pattern matrix for each role to capture its compatibility with all involved entities. 
It further explores the multilinear scoring function with efficient computation achieved. 
We show that RAM is fully expressive, 
which also provides a unified view of most bilinear models in binary relational KBs. 
The key insight of RAM is to model n-ary relational KBs from the role level, 
with the sharing information considered. 
Our contributions are as follows:
\begin{itemize}[leftmargin=*]
	\item We propose RAM for n-ary relational KBs, which achieves linear complexity in both time and space. 
	Especially, RAM identifies the importance of roles in n-ary relations, and learns latent space as well as pattern matrices for roles, to capture semantic relatedness and role-entity compatibility, respectively.
	\item We prove that RAM is fully expressive to represent any given n-ary relational KBs, and also demonstrate that RAM generalizes several existing binary relational KB modeling approaches.
	\item Extensive experiments on five benchmark datasets show that RAM achieves state-of-the-art performance on n-ary relational knowledge base completion and comparable performance on binary relational KBs. Further, several visualization results demonstrate that RAM successfully captures the role semantics.
\end{itemize}

\section{Related Work}\label{sec:related work}
\subsection{Binary Relational KB Modeling} 
As aforementioned, 
existing approaches in binary relational KBs focus on the relation level modeling (without roles considered), 
which fall into three categories: 
translational models, neural network models and bilinear models \citep{zhangautosf,rossi2020knowledge}.

Typical translational models of TransE \citep{bordes2013translating}, TransH \citep{TransH}, 
and TransR \citep{TransR} project entities into a latent space via relation operations, 
and measure the fact plausibility by distance metrics therein \citep{wang2017knowledge,liu2020generalizing}, 
which are less expressive to represent some relation patterns \citep{kazemi2018simple,sun2019rotate}. 
On the other hand, 
the neural network models rely on various network block designs, 
such as the convolutional neural network (CNN) in ConvE \citep{ConvE}, 
and graph convolutional neural network (GCN) in R-GCN \citep{schlichtkrull2018modeling} and CompGCN \citep{vashishth2019composition}. 
Such models own tremendous parameters for expressiveness, 
which lead to intractable training and overfitting \citep{kazemi2018simple,liu2020generalizing}. 
Especially, bilinear models are the most relevant work to ours. 
The bilinear models define the scoring function as a bilinear product of entity/relation embeddings, 
and introduce several customized designs for expressiveness, 
such as equivalent embeddings in DistMult \cite{yang2014embedding}, 
complex-valued embeddings in ComplEx \cite{trouillon2016complex}, 
quaternion embeddings in QuatE \cite{zhang2019quaternion}, 
inverse relations in SimplE \cite{kazemi2018simple}, 
core tensor in TuckER \cite{balazevic2019tucker}, 
and automated relation matrix in AutoSF \cite{zhangautosf}. 

However, above models are limited in binary relational modeling with several binary relation-constrained designs. 
Considering the complicated role semantics in n-ary relations, 
it is none-trivial to extend such models to the n-ary case, 
where multiple entities involve with n-ary relations \citep{liu2020generalizing}. 
Besides, some binary relational approaches \citep{xie2016representation,krompass2015type} incorporate entity/relation types into KB modeling. 
For example, in Figure~\ref{fig:example}, 
the types of \textit{Schwarzenegger} and \textit{California} are Person and Location, respectively. 
Such type information focuses on the entity/relation itself, 
far from our proposed role, which emphasizes the semantics of entities to the relation.

\begin{table*}[ht]
	\caption{
		A comparison of n-ary relational KB modeling approaches, 
		The role-aware property considers the semantic relatedness among roles and role-entity compatibility. 
		N/A terms are not definite in literature. 
		$n_e$ and $n_r$ denote number of entities and relations respectively. $n_a$ is maximum arity in KB and $d$ is embedding dimensionality. $\bm{e}_i$ is the embedding of entity $e_i$ and $\bm{r}$ is the embedding of relation $r$. $\bm{a}^r$ and $\bm{w}^r$ are relation-dependent vectors. $K<d\ll n_r$ is latent space size. $\text{FCN}$ denotes fully connected networks, $\text{vec}(\cdot)$ is vectorization operation and $\left[\cdot,\cdot\right]$ is vector concatenation. $\min(\cdot)$ is element-wise minimizing operation, $*$ is convolution operator and $\Omega$ is convolutional filter. }
	\label{tab:MultiAry_SF}
	\centering
	\vspace{-10px}
	\setlength\tabcolsep{4pt}
	\begin{tabular}{c c c c c c}
		\toprule
		\textbf{Model} & \textbf{$\!\!\!\!\!\!$Role-aware} & \textbf{Expressive} & \textbf{Scoring Function} & $\bm{\mathcal{O}}_{\text{time}}$ &
		{$\bm{\mathcal{O}}_{\text{space}}$}  
		\\
		
		\midrule
		m-TransH \citep{wen2016representation}
		& \XSolidBrush
		& \XSolidBrush
		& {$\|\sum^{a_r}_{i=1}\bm{a}^{r}[i]\cdot\left(\bm{e}_i-{\bm{w}^r}^\top\bm{e}_i{\bm{w}^r}\right)\!+\!\bm{r}\|^2$}
		& $\mathcal{O}(d)$
		& $\mathcal{O}\left(n_ed\!+\!2n_rd\right)$ \\
		
		\midrule
		RAE \citep{zhang2018scalable} 
		& \XSolidBrush
		& \XSolidBrush
		& {$\|\sum^{a_r}_{i=1}\bm{a}^{r}[i]\cdot\left(\bm{e}_i-{\bm{w}^{r}}^\top\bm{e}_i{\bm{w}^r}\right)\!+\!\bm{r}\|^2\!+\!\lambda\sum_{i,j} \scriptsize{\text{FCN}}\left(\left[\bm{e}_i,\bm{e}_j\right]\right)$}
		& $\mathcal{O}(d^2)$
		& $\mathcal{O}\left(n_ed\!+\!2n_rd\right)$ 
		\\
		
		\midrule
		NaLP \citep{guan2019link}  
		& \XSolidBrush
		& N/A
		& ${\scriptsize\text{FCN}}_2 ( \min\limits_{i,j} ({\scriptsize\text{FCN}}_1 ( 
		\bigl[ [\bm{u}^{r}_{i},\bm{e}_i] \! * \! \bm{\Omega}, [\bm{u}^{r}_{j},\bm{e}_j] \! * \! \bm{\Omega}  \bigr]
		) ) )$
		& $\mathcal{O}(d^2)$
		& $\mathcal{O}\left(n_ed\!+\!n_rn_ad\right)$ \\
		
		\midrule
		HINGE \citep{rossobeyond} 
		& \XSolidBrush 
		& N/A
		& {${\scriptsize\text{FCN}} ( \min\limits_i ( \bigl[{\small\text{vec}} ([\bm{r},\bm{e}_1,\bm{e}_2]*\bm{\Omega}_1), {\small\text{vec}} ([\bm{r},\bm{e}_1,\bm{e}_2, \bm{u}^{r}_{i}, \bm{e}_i]*\bm{\Omega}_2) \bigr]) )$}
		& $\mathcal{O}(d^2)$
		& $\mathcal{O}\left(n_ed\!+\!n_rn_ad\right)$ 	\\
		
		\midrule
		NeuInfer \citep{guan2020neuinfer}  
		& \XSolidBrush
		& N/A	
		& {$\alpha\cdot{\scriptsize\text{FCN}}_1 ([\bm{r},\bm{e}_1,\bm{e}_2]) + (1-\alpha)\cdot {\scriptsize\text{FCN}}_3 ( \min\limits_{i} ({\scriptsize\text{FCN}}_2 ([\bm{r},\bm{e}_1,\bm{e}_2, \bm{u}^{r}_{i}, \bm{e}_i]) ) ) $}
		& $\mathcal{O}(d^2)$
		& $\mathcal{O}\left(n_ed\!+\!n_rn_ad \right)$  \\
		
		\midrule
		HypE \citep{fatemi2019knowledge} 
		& \XSolidBrush
		& \Checkmark 
		& {$\langle \bm{r}, \bm{e}_1*\bm{\Omega}_1,\cdots,  \bm{e}_{a_r}*\bm{\Omega}_{a_r} \rangle$} 
		& $\mathcal{O}(d)$
		& $\mathcal{O}\left(n_ed\!+\!n_rd\right)$ \\
		
		\midrule
		RAM (ours) 
		& \Checkmark
		& \Checkmark 
		& $\sum_{i=1}^{a_r}  \langle \bm{u}^{r}_{i}, {{\bm{P}}^{r}_{i}{[1,:]}}\bm{E}_1, \cdots, {\bm{P}}^{r}_{i}[a_r,:]\bm{E}_{a_r} \rangle$
		& $\mathcal{O}(d)$
		& $\mathcal{O}\left(n_ed\!+\!Kn_rn_a\right)$	\\
		\bottomrule
	\end{tabular}
\end{table*} 

\subsection{N-ary Relational KB Modeling}
The early models of m-TransH \cite{wen2016representation} and RAE \cite{zhang2018scalable} are generalized from TransH \cite{TransH}, 
where the weighted sum of projected entities returns the plausibility score. 
RAE further considers the relatedness of involved entities with fully connected networks (FCNs).  
However, above models still keep the weak expressiveness of TransH \cite{kadlec2017knowledge}, 
thus obtain relatively weak performance in practice. 

The multilinear models extend bilinear models via the multilinear product. 
For instance, HypE \cite{fatemi2019knowledge} generalizes SimplE \citep{kazemi2018simple}, utilizes convolutional filters for entity embeddings, and multilinear product for plausibility measure. 
Another model GETD \cite{liu2020generalizing} generalizes TuckER \cite{balazevic2019tucker} to the n-ary case with tensor ring decomposition. 
However, GETD can only apply in KBs with single-arity relations \citep{StarE}. 
Furthermore, 
both existing translational and multilinear models are in relation level modeling, 
completely missing the role semantics in n-ary relations.

In contrast, the neural network model NaLP \cite{guan2019link} introduces the role-entity form in n-ary relational KBs, 
where CNNs and FCNs are utilized to measure the compatibility between the role and its mapping entity. 
Its following works of HINGE \cite{rossobeyond} and NeuInfer \cite{guan2020neuinfer} decompose the n-ary relational fact into a triplet and several role-entity pairs. 
Especially, HINGE mainly captures the compatibility with CNNs while NeuInfer only relies on FCNs. 
Similarly, a recent work StarE \cite{StarE} firstly applies CompGCN to model the decomposed triplets, 
however, fails to predict the missing role-entity pairs, 
i.e., StarE only focuses on modeling part of n-ary relational facts. 
In these models, each role is learned independently, 
without considering the semantic relatedness among roles. 
Moreover, for a role, only the compatibility with its mapping entity is considered, 
losing the whole compatibility with other involved entities. 
Besides, neural network models leverage too many parameters for expressiveness, 
which are prone to overfitting and make training intractable \cite{liu2020generalizing,StarE,kazemi2018simple}. 

Hence, most existing n-ary relational KB modeling approaches are extended from binary relational approaches, 
which are summarized in Table~\ref{tab:MultiAry_SF}. 
Compared with existing n-ary relational modeling approaches, 
the proposed RAM carefully models the role semantics in n-ary relational KBs, 
with full expressiveness as well as linear time and space complexity achieved.

\section{Method}\label{sec:model}

We begin by describing the n-ary relational KB modeling problem and introducing notation. 

\subsection{Problem Setup}\label{subsec:problem}

According to the n-ary relation definition in \citep{giunti2019representing,guan2019link,noy2006defining,rouces2015framebase}, 
in an n-ary relational KB $\mathcal{B}=(\mathcal{E}, \mathcal{R}, \mathcal{F})$ 
with entity set $\mathcal{E}$, 
relation set $\mathcal{R}$ and 
observed fact set $\mathcal{F}$, 
for relation $r\in\mathcal{R}$ with arity $a_r$ \citep{fatemi2019knowledge}, 
its semantic roles are defined as, 
\begin{definition}[Role]
	The $a_r$ roles, $(\gamma^r_1,\cdots,\gamma^{r}_{a_r})$, form the relation $r$,	
	which identify semantics of involved entities to  $r$.
\end{definition}

The roles can be explicit or implicit, 
which will be 
discussed in Remark~\ref{remark:explicit}. 
Hence, an n-ary relational fact in $\mathcal{F}$ is expressed in role-entity form $\{\gamma^{r}_{1}:e_1,\cdots,\gamma^{r}_{{a_r}}:e_{a_r}\}$ with $e_{i}\in\mathcal{E}$ and $i = 1,\cdots,a_r$. 
To handle the semantic relatedness as well as compatibility with roles, 
the n-ary relational KB modeling can be specified as role-aware n-ary relational knowledge base completion (KBC) problem, 
which is defined as follows.

\begin{problem}[Role-aware n-ary relational KBC]
	Given an incomplete n-ary relational KB $\mathcal{B}=(\mathcal{E}, \mathcal{R}, \mathcal{F})$, 
	the role-aware n-ary relational KBC problem stresses the semantic roles existed in n-ary relations, exploiting the existing facts $\mathcal{F}$ to infer missing ones. 
\end{problem}

In the sequel,
we denote scalars by lowercase letters, vectors by bold lowercase letters and matrices by bold upper case letters. For indexing, we denote $\bm{a}[i]$ as the $i$-th element of a vector $\bm{a}$, $\bm{A}[i,j]$ as the $[i,j]$-th element of matrix $\bm{A}$, and $\bm{A}[i,:]$ as the $i$-th row of $\bm{A}$. Besides, $\left\langle\cdot\right\rangle$ represents the multilinear product, written as $\left\langle\bm{a}_{1},\bm{a}_{2},\cdots,\bm{a}_{n}\right\rangle=\sum_i \bm{a}_{1}[i]\bm{a}_{2}[i]  \cdots\bm{a}_{n}[i]$.
The related notations frequently used in this paper are listed in Table~\ref{tab:notation}.

\begin{table}[t]
	\caption{List of key symbols. }
	\label{tab:notation}
	\centering
	\vspace{-10px}
	\begin{tabular}{c | c}
		\toprule
		Symbol & Meaning\\
		\hline
		$\mathcal{B}=(\mathcal{E},\mathcal{R},\mathcal{F})$ & n-ary relational KB \\
		$\mathcal{E}$ & set of entities \\
		$\mathcal{R}$ & set of relations \\
		$\mathcal{F}$ & set of observed facts \\
		$a_r$ & the arity of relation $r$ \citep{fatemi2019knowledge} \\
		$\hat{\bm{u}}_k$ & $k$-th basis vector in latent space for roles \\
		$\bm{u}^r_i$ & $i$-th role representation of relation $r$  \\
 		$\hat{\bm{P}}_k$ & $k$-th basis matrix \\
 		$\bm{P}^r_i$ & pattern matrix for $i$-th role of relation $r$  \\
		$\bm{\alpha}^r_i$ & weight vector for $i$-th role of relation $r$\\
		$\Phi(\cdot)$ & element-wise softmax function \\
		$\langle\cdot\rangle$ & multilinear product\\
		\bottomrule
	\end{tabular}
\end{table} 
\noindent

\subsection{Role-Aware Modeling Design}\label{sec:model design}
As discussed before, 
the roles are the important element of n-ary relational KBs, 
which identify the semantics of entities to relations. 
Especially, such the roles further determine the fact plausibility in KB modeling from the following two aspects. 
\begin{itemize}[leftmargin=*]
	\item First, several roles are semantically related and shared in n-ary relational facts. 
	A key observation on n-ary relational dataset WikiPeople \citep{guan2019link} is that over 80\% roles in beyond-ternary relations also appear in lower-arity relations. 
	\item Second, the compatibility between the role and all involved entities contributes to fact plausibility, 
	e.g., 
	in Figure~\ref{fig:example} 
	the role \textsc{Movie} with its mapping entity \textit{Terminator 2} and involved entities of \textit{Schwarzenegger} and \textit{T-800}  affect the plausibility together. 
	The compatibility relates to knowledge in KBs.
\end{itemize}

Therefore, 
our proposed approach RAM models n-ary relational KBs from the essential role level, 
which leverages a latent space for semantic relatedness among roles, 
and a pattern matrix for compatibility capture between the role and all involved entities. 
Finally, the multilinear product is adopted for plausibility measure, 
which achieves both full expressiveness and linear complexity.


\subsubsection{Latent space for roles} 
The semantic relatedness among roles in n-ary relational facts are just like the sharing information in several machine learning areas, 
i.e., 
the topic sharing across multiple documents in topic modeling \citep{teh2005sharing}, 
the task relatedness across multiple tasks in multi-task learning \citep{argyriou2007multi}, 
as well as the sparse representations across multiple signals in sparse coding \citep{aharon2006k}. 
Inspired by such success, 
to fully exploit the potential of shared roles,
we build a latent space for roles with $K$ latent basis vectors $\hat{\bm{u}}_{i}\in\mathbb{R}^d$ for $i = 1,\cdots,K$ ($K\ll n_r$). 
The embedding of each role is computed based on a combination of basis vectors as follows,
\begin{align}
	\bm{u}^{r}_{i}
	= \sum\nolimits^{K}_{k=1}\Phi(\bm{\alpha}^{r}_{i})[k]\cdot \hat{\bm{u}}_k, 
	\quad \forall
	i = 1,2,\cdots,a_r, \label{eq:role-embedding}
\end{align}
where
$\bm{u}^{r}_{i}$ is embedding vector for the $i$-th role of relation $r\in\mathcal{R}$, 
and $\bm{\alpha}^{r}_{i}\in\mathbb{R}^K$ is the corresponding weight vector, referred to as role weight. 
In this way, the semantic relatedness is implicitly parameterized by the role weight. 
It is further normalized by the element-wise softmax function $\Phi$ as follows,
\begin{align*}
\Phi(\bm{\alpha}^{r}_{i})[k]
= e^{\bm{\alpha}^{r}_{i}{[k]}} 
/ \sum\nolimits_{k'=1}^{K} e^{\bm{\alpha}^{r}_{i}{[k']}},
\quad \forall k=1,2,\cdots,K. 
\end{align*}

As for the entity, 
since the entity might own multiple semantics like the \emph{star} and the \emph{political} semantics of \textit{Schwarzenegger} in Figure~\ref{fig:example}, 
we design a multi-embedding mechanism \citep{tran2019analyzing}, 
and map each entity $e_i\in\mathcal{E}$ to multiple ($m$) embeddings for semantics, 
denoted by $\bm{E}_i\in\mathbb{R}^{m\times d}$, 
where $d$ is the embedding dimensionality.

\begin{figure}[ht]
	\centering
	\includegraphics[width=0.75\linewidth,height=50mm]{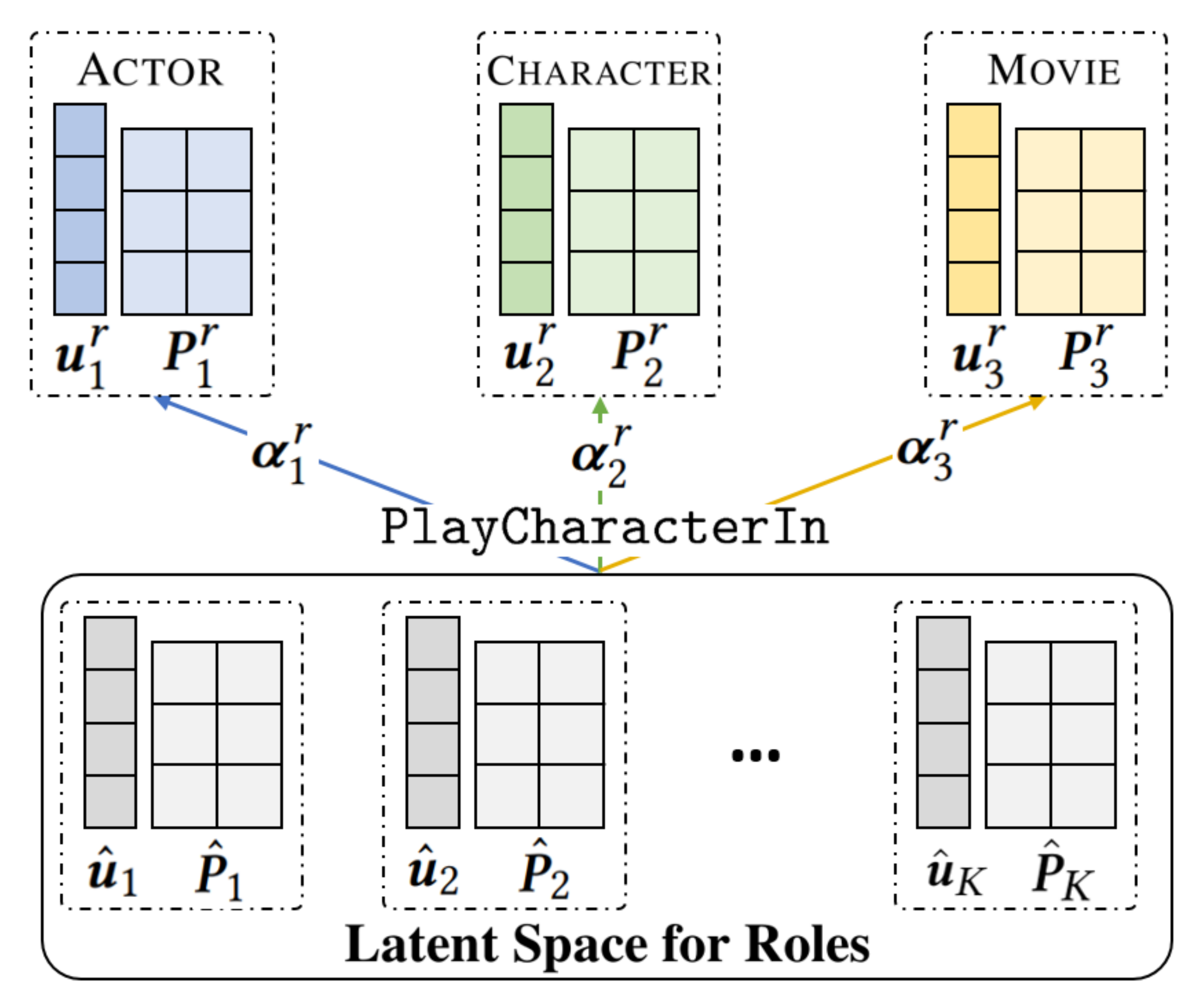}
	\vspace{-10px}
	\captionof{figure}{Latent space for roles and pattern matrices in RAM, with the multiplicity of entity embeddings $m=2$ (in example of ternary relation \texttt{PlayCharacterIn} with roles (\textsc{Actor}, \textsc{Character},  \textsc{Movie})).}
	\label{fig:architecture}
\end{figure}

\subsubsection{Pattern matrix}
To measure the compatibility between the role and all involved entities, 
each role is learned with a role-aware pattern matrix, 
i.e., for a relation $r\in\mathcal{R}$, 
the pattern matrix of the $i$-th role is denoted by $\bm{P}^{r}_{i}\in\mathbb{R}^{a_r\times m}$, 
where the $j$-th row $\bm{P}^{r}_{i}[j,:]$ indicates the compatibility with multiple embeddings of the $j$-th entity involved in facts. 
With the designed latent space for roles, 
the pattern matrix can be learned as follows,
\begin{align}
\bm{P}^{r}_{i}
=
\sum\nolimits^{K}_{k=1}\Phi(\bm{\alpha}^{r}_{i})[k]\cdot \Phi(\hat{\bm{P}}_k),  
\quad \forall 
i = 1,2,\cdots,a_r, 
\label{eq:pattern-matrix}
\end{align}
where $\hat{\bm{P}}_k\in\mathbb{R}^{a_r\times m}$ is the basis matrix\footnote{Each basis vector $\hat{\bm{u}}_k$ actually links with a $\hat{\bm{P}}_{k,2}\in\mathbb{R}^{2\times m}$ for binary relations, a $\hat{\bm{P}}_{k,3}\in\mathbb{R}^{3\times m}$ for ternary relations, etc. Here we use a $\hat{\bm{P}}_{k}\in\mathbb{R}^{a_r\times m}$ for simplicity.} 
linked with basis vector $\hat{\bm{u}}_k$ in latent space. 
The entire basis matrix is also normalized by $\Phi$. 
Figure~\ref{fig:architecture} illustrates the latent space for roles. 
The basis pattern matrix $\hat{\bm{P}}_k$ is aligned with the latent basis vector $\hat{\bm{u}}_k$, 
which are used to compute for role embeddings and role-aware pattern matrices.

\subsubsection{Multilinear scoring function} 
For effective and efficient performance, we calculate the fact plausibility in a multilinear product way, 
which introduces few parameters for light and easy training. 
For each true fact $x\coloneqq\{\gamma^{r}_{1}:e_1,\cdots,\gamma^{r}_{{a_r}}:e_{a_r}\}$, 
the obtained scoring function is as follows,
\begin{align}
\phi(x) =  
\sum\nolimits_{i=1}^{a_r} 
\left\langle 
\bm{u}^{r}_{i}, {{\bm{P}}^{r}_{i}{[1,:]}}\bm{E}_1, \cdots, {\bm{P}}^{r}_{i}[a_r,:]\bm{E}_{a_r} 
\right\rangle, 
\label{eq:score}
\end{align}
where ${\bm{P}}^{r}_{i}{[1,:]}\bm{E}_1$ captures the compatibility between the $i$-th role $\gamma^r_i$ and the first involved entity $e_i$, 
i.e., the multiple embeddings of $e_i$ is weighted by the elements of ${\bm{P}}^{r}_{i}{[1,:]}$. 
Besides, each summation term in multilinear product form is the compatibility of corresponding role with all involved entities . 


\begin{remark}\label{remark:explicit}
RAM applies to n-ary relational KBs with explicit or implicit roles. 
Specifically, 
some n-ary relational KB datasets like WikiPeople \citep{guan2019link} provide the explicit role-entity form, 
where different relations explicitly share roles, 
e.g., role \textsc{Movie} in relation \texttt{PlayCharacterIn} and \texttt{DirectorAwardFor} in Figure~\ref{fig:example}. 
On the other hand, 
datasets like JF17K \citep{wen2016representation} and FB-AUTO \cite{fatemi2019knowledge} only provide the n-ary relation and entities, 
where the semantic roles are implicitly shared across relations. 
Since RAM relies on latent space for roles, 
either explicit or implicit roles can be parameterized by role weights in RAM.
\end{remark}

\subsection{Stochastic Model Training}
With the scoring function obtained above,
we design the training loss as well as the learning objective. 
Usually, KBs only provide positive (true) observations, 
while the negative ones for training are not available. 
Thus, we develop a negative sampling strategy for the n-ary relational KB modeling.
For each true fact $x\in\mathcal{F}$, we obtain the negative samples as:
\begin{align*}  
	\bigcup^{a_r}_{i=1} \mathcal{N}^{(i)}_{x}
	\equiv
	\bigcup^{a_r}_{i=1}
	\left\lbrace 
	\gamma^{r}_{1}\!:\!e_1,\cdots, \gamma^{r}_{i}\!:\!\bar{e}_i,\cdots,\gamma^{r}_{{a_r}}\!:\!e_{a_r}\notin \mathcal{F}|\bar{e}_i\in \mathcal{E}, \bar{e}_i \neq e_i
	\right\rbrace,
\end{align*}
where $\mathcal{N}^{(i)}_x$ replaces the entity linked with $\gamma^{r}_{i}$. 
The strategy generalizes from the ones in the binary case \citep{lacroix2018canonical}. 
Furthermore, we adopt the instantaneous multi-class log-loss \citep{lacroix2018canonical,liu2020generalizing}, 
and formulate a minimizer of an empirical risk as follows,
\begin{align}
	\min\limits_{\substack{
			\{\bm{E}_i\},
			\{\hat{\bm{u}}_{i}\},\\	
			\{\hat{\bm{P}}_{i}\},
			\{\bm{\alpha}^{r}_{i}\}}}
	\sum\nolimits_{x\in\mathcal{F}} \!\! \sum\nolimits^{a_r}_{i=1}\!\! -\log
	\left[ 
	e^{\phi(x)}\!\! /\! \big( e^{\phi(x)}
	\!\!+\!\!
	\sum\nolimits_{y\in\mathcal{N}^{(i)}_{x}} e^{\phi(y)}
	\big) 
	\right]
	\label{eq:obj},
\end{align}
where the softmax form of loss guarantees that exactly one correct sample is learned among the candidates \citep{kadlec2017knowledge}. 

Algorithm~\ref{alg:training} summarizes the training procedure for RAM. 
For each sampled fact in the n-ary relational KB, 
we first obtain the negative samples, 
as well as the embedding of roles and pattern matrices from latent space. 
Then, we calculate the plausibility score for the sample. 
Finally, RAM is trained in a mini-batch way to minimize the empirical risk formulation above.

\SetKwInput{KwInit}{Init}
\begin{algorithm}[t]
	\SetAlgoNoLine
	\caption{Training procedure for RAM} \label{alg:training}
	\KwIn{N-ary relational KB $\mathcal{B}=(\mathcal{E},\mathcal{R},\mathcal{F})$, the latent space size $K$}
	\KwOut{Entity and role embeddings, pattern matrices}
	\KwInit{{$\bm{E}$ for $e\in\mathcal{E}$, $\{\bm{\alpha}^{r}_{i}\}_{i=1}^{a_r}$ for $r\in\mathcal{R}$, $\{\hat{\bm{u}}_k\}^K_{k=1}$ and $\{\hat{\bm{P}}_k\}_{k=1}^K$}}
	
	\For{$t=1,\cdots,n_\textnormal{iter}$}{
		Sample a mini-batch $\mathcal{F}_{\text{batch}}\in \mathcal{F}$ of size $m_b$; 
		\\
		
		\For{ \textnormal{each} $x\coloneqq\{\gamma^{r}_{1}:e_1,\cdots,\gamma^{r}_{a_r}:e_{a_r}\}\in\mathcal{F}_{\textnormal{batch}}$}
		{
			\raggedright{Construct negative samples for fact $x$;}
			\\
			
			\raggedright{$\bm{u}^{r}_{i}\leftarrow$ compute role embeddings using \eqref{eq:role-embedding};} 
			\\
			
			\raggedright{$\bm{P}^{r}_{i}\leftarrow$ compute pattern matrices using \eqref{eq:pattern-matrix}; }
			\\
			
			
		}
		\raggedright{Update learnable parameters w.r.t. gradients based on the whole objective in \eqref{eq:obj};}
		
	}
\end{algorithm}

\section{Theoretical Understanding} 
\label{sec:theoretical}
In this section, 
we show that RAM carefully learns from binary relational KB modeling, 
which achieves the linear complexity and full expressiveness. 
Moreover, RAM unifies the scoring functions of several bilinear models.

\begin{table*}[ht]
	\caption{Dataset statistics. "Arity" denotes the involved arities of relations. 
		"\#$\geq$5-ary" denotes the amount of facts with 5-ary relations and beyond.}
	\label{tab:datasets}
	\vspace{-10px}
	\centering
	\begin{tabular}{ccccccccccc}
		\toprule
		Dataset               & $\vert\mathcal{E}\vert$ & $\vert\mathcal{R}\vert$ & Arity & \#Train & \#Valid & \#Test & \#2-ary & \#3-ary & \#4-ary & \#$\geq$5-ary \\ \midrule
		WikiPeople \citep{guan2019link}   &         47,765          &           707           &  2-9  & 305,725 & 38,223  & 38,281 & 337,914 & 25,820  & 15,188  &     3,307     \\
		JF17K \citep{zhang2018scalable}   &         28,645          &           322           &  2-6  & 61,104  & 15,275  & 24,568 & 54,627  & 34,544  &  9,509  &     2,267     \\
		FB-AUTO \citep{fatemi2019knowledge} &          3,388          &            8            & 2,4,5 &  6,778  &  2,255  & 2,180  &  3,786  &    0    &   215   &     7,212     \\ \midrule
		WN18 \citep{bordes2013translating}  &         40,943          &           18            &   2   & 141,442 &  5,000  & 5,000  & 151,442 &    -    &    -    &       -       \\
		FB15k \citep{bordes2013translating} &         14,951          &          1,345          &   2   & 484,142 & 50,000  & 59,017 & 593,159 &    -    &    -    &       -       \\ \bottomrule
	\end{tabular}
\end{table*}

\subsection{Complexity Analysis}\label{sec:comp}
Table~\ref{tab:MultiAry_SF} summarizes the role-aware property, expressiveness, as well as scoring functions of existing n-ary relational approaches, 
in terms of time and space complexity. 
Due to the efficiency of the multilinear product,
our RAM achieves the linear time complexity of $\mathcal{O}(d)$. 
On the other hand, for an n-ary relational KB with the maximum arity of $n_a$, 
the latent space with $\{\hat{\bm{u}}_{i}\}$ and $\{\hat{\bm{P}}_{i}\}$ costs at most $\mathcal{O}(Kd+Kmn_a)$ parameters, 
while role weights $
\{\bm{\alpha}^{r}_{i}\}$ cost $\mathcal{O}(Kn_rn_a)$ parameters. 
As shown in dataset statistics in Table~\ref{tab:datasets}, 
the arity of over 95\% relations are smaller than 4, and $m$ is generally assigned with small values like 2. 
Hence the space complexity is $\mathcal{O}(n_ed+Kn_rn_a+Kd+Kmn_a)=\mathcal{O}(n_ed+Kn_rn_a)$ ($K\ll n_r$). 
Thus, our approach is scalable to large datasets as it remains linear in both time and space \citep{nickel2015review}.

\subsection{Expressive Power of RAM}\label{sec:expressive}

Full expressiveness is another important property for learning capacity measure \citep{balazevic2019tucker,fatemi2019knowledge,kazemi2018simple,rossi2020knowledge,trouillon2016complex,liu2020generalizing}. 
Specifically, a model is fully expressive if given any ground truth over facts in KB, 
there exists at least one assignment of embedding values for the model that correctly separates valid facts from invalid facts.
We theoretically establish the full expressiveness of RAM with embedding dimensionality bound presented in Theorem~\ref{theorem:full expressive}.

\begin{theorem}\label{theorem:full expressive}
	For any ground truth over entities $\mathcal{E}$ and relations $\mathcal{R}$ of the n-ary relational KB containing $\eta\geq1$ true facts, there exists a RAM model with the embedding dimensionality $d=\eta$, the multiplicity of entity embedding $m=\max_{r\in\mathcal{R}} a_r$, and the latent space size $K=\eta$ that accurately represents that ground truth.
\end{theorem}

The proof is in Appendix \ref{app:full expressive}. 
Theorem~\ref{theorem:full expressive} demonstrates the fully expressive RAM model, 
which has the potential to learn correctly any valid n-ary relational KBs. 
In contrast, translational models like m-TransH \citep{wen2016representation} and RAE \citep{zhang2018scalable} have restrictions on represented relations \citep{kazemi2018simple}, 
and thus are not fully expressive, 
while neural network models like NaLP \cite{guan2019link}, HINGE \citep{rossobeyond} and NeuInfer \cite{guan2020neuinfer} approximate full expressiveness with tremendous parameters. Based on Occam's Razor \citep{blumer1987occam}, 
RAM with linear complexity is much more powerful than most existing n-ary relational models in Table~\ref{tab:MultiAry_SF}. 
This also guarantees the performance of RAM in binary relational KBs, representing various types of relational patterns including the symmetric pattern, inverse pattern, etc.  

\begin{table*}[ht]
	\caption{N-ary relational KBC results. Best results are in bold and second best results are underlined. Results of NaLP and NeuInfer on WikiPeople and results of HypE on FB-AUTO are copied from original papers.}\label{tab:resuls_n-ary}
	\centering
	\vspace{-10px}
	\begin{tabular}{ccccc|cccc|cccc}
		\toprule
		&\multicolumn{4}{c}{\textbf{WikiPeople}} & \multicolumn{4}{c}{\textbf{JF17K}} & \multicolumn{4}{c}{\textbf{FB-AUTO}}\\
		\textbf{Model} & {MRR} & {Hit@10} & {Hit@3}& {Hit@1} & {MRR} & {Hit@10} & {Hit@3} & {Hit@1} & {MRR} & {Hit@10}& {Hit@3} & {Hit@1}  \\
		\midrule
		RAE \citep{zhang2018scalable}  & {0.253} & {0.463} & 0.343  & 0.118 & {0.396} & {0.561} & 0.433 & 0.312 & {0.703}  & {0.854}& 0.764 & 0.614 \\
		NaLP \citep{guan2019link}  & {0.338} & {0.466} & 0.364 & {0.272} & {0.310} & {0.450} & 	0.334 & 0.239 & {0.672} & {0.774}& 	0.712 & 0.611\\
		HINGE \citep{rossobeyond} & {0.333} & {0.477}& 0.361 & {0.259} & {0.473}  & {0.618}&	0.490  & 0.397 & {0.678} & {0.765}& 0.706 & 0.630\\
		NeuInfer \citep{guan2020neuinfer} & {0.350} & {0.467}& 0.381  & {0.282} &{0.451} & {0.604}& 0.484 & {0.373} & {0.737} & {0.805 }& 0.755  & {0.700} \\
		HypE \citep{fatemi2019knowledge}  & {0.292} & {0.502}& 0.375 & 0.162 &\underline{0.507} & \underline{0.669}& \underline{0.550} & \underline{0.421} & \underline{0.804} & \underline{0.856}& \underline{0.824} & \underline{0.774} \\
		\midrule
		{RAM/e} & {\underline{0.370}} & { \underline{0.507}} & \underline{0.410} & \textbf{0.293} & - &- & - & -& -& - & - & -\\
		{RAM} & {\textbf{0.380}} & { \textbf{0.539}} & \textbf{0.445} & \underline{0.279} & { \textbf{0.539}} & { \textbf{0.690}} & \textbf{0.573} & \textbf{0.463} & { \textbf{0.830}} & { \textbf{0.876}} & \textbf{0.851} & \textbf{0.803}\\
		\bottomrule
	\end{tabular}
\end{table*}

\subsection{Connection to Existing Bilinear Models}\label{sec:conblm}
In Proposition~\ref{theorem:conblm}, 
we demonstrate that RAM generalizes several bilinear models in binary relational KBs. 
This also explains its strong empirical performance compared to other baselines in experiments.
\begin{prop} \label{theorem:conblm}
	The scoring functions of DistMult \cite{yang2014embedding} and SimplE \cite{kazemi2018simple} are the special case of RAM in binary relational KBs, by assigning different pattern matrices and multiple embeddings. The scoring functions of ComplEx \citep{trouillon2016complex} and QuatE \cite{zhang2019quaternion} can also be generalized by appropriate modification to RAM.
\end{prop}
Here, we illustrate Proposition~\ref{theorem:conblm} with DistMult and SimplE, 
and leave ComplEx and QuatE in Appendix~\ref{app:gblm}. 
Specifically, each entity is assigned with two embeddings, e.g., $\bm{E}_h=[\bm{e}_{h,1};\bm{e}_{h,2}]$, and the binary relation $r$ corresponds to the role embeddings of $\bm{u}^r_1$ and $\bm{u}^r_2$. With specific pattern matrices, RAM rewrites scoring functions as:
\begin{align*} 
\text{DistMult: }
& \langle\bm{u}^r_{1},\frac{1}{2}\bm{e}_{h,1},\frac{1}{2}\bm{e}_{t,1}\rangle\!+\!\langle \bm{u}^r_{2},\frac{1}{2}\bm{e}_{h,2},\frac{1}{2}\bm{e}_{t,2}\rangle \notag\\
&=\!\langle \bm{u}^r_{1},\bm{P}^{r,\text{DM}}_{1}[1,:]\bm{E}_h,\bm{P}^{r,\text{DM}}_{1}[2,:]\bm{E}_t \rangle\! \notag \\
&+
\!\langle \bm{u}^r_{2},\bm{P}^{r,\text{DM}}_{2}[1,:]\bm{E}_h,\bm{P}^{r,\text{DM}}_{2}[2,:]\bm{E}_t \rangle,  \notag
\\
\text{where} \;& \bm{P}_{1}^{r,\text{DM}}={
	\left[ \arraycolsep=3pt
	\begin{array}{cc} 
		\nicefrac{1}{2} & 0 \\[3pt]
		\nicefrac{1}{2} & 0	
	\end{array} 
	\right ]},
\bm{P}_{2}^{r,\text{DM}}={
	\left[ \arraycolsep=3pt
	\begin{array}{cc}
		0 & \nicefrac{1}{2} \\[3pt]
		0 & \nicefrac{1}{2} 	
	\end{array} 
	\right ]},\notag\\
\text{SimplE: }
&\langle\bm{u}^{r}_{1},\frac{1}{2}\bm{e}_{h,1},\frac{1}{2}\bm{e}_{t,2}\rangle\!+\!\langle \bm{u}^{r}_{2},\frac{1}{2}\bm{e}_{h,2},\frac{1}{2}\bm{e}_{t,1}\rangle \notag\\
&=\!\langle \bm{u}^{r}_{1},\bm{P}^{r,\text{SE}}_{1}[1,:]\bm{E}_h,\bm{P}^{r,\text{SE}}_{1}[2,:]\bm{E}_t \rangle\!  \notag \\
&+\!\langle \bm{u}^{r}_{2},\bm{P}^{r,\text{SE}}_{2}[1,:]\bm{E}_h,\bm{P}^{r,\text{SE}}_{2}[2,:]\bm{E}_t \rangle,  \notag
\\
\text{where} \;&
\bm{P}_{1}^{r,\text{SE}}={
	\left[ \arraycolsep=3pt
	\begin{array}{cc}
		\nicefrac{1}{2} & 0 \\[3pt]
		0 & \nicefrac{1}{2}	
	\end{array} 
	\right ]},
\bm{P}_{2}^{r,\text{SE}}={
	\left[ \arraycolsep=3pt
	\begin{array}{cc}
		0 & \nicefrac{1}{2} \\[3pt]
		\nicefrac{1}{2} & 0	
	\end{array} 
	\right ]},\!\!\!\notag
\end{align*}
For DistMult, its equivalent embeddding design can be considered as pattern matrices with only one non-zero column. 
For SimplE, its inverse relation design can be viewed as diagonal/anti-diagonal pattern matrices. Note that the pattern matrices are predefined and role-independent in bilinear models, 
while learnable and role-aware in RAM. Thus, RAM provides a unified view of bilinear models with a more flexible design.

\section{Experiments and Results}\label{sec:exp}
In this section, we evaluate RAM on standard KBC tasks with both n-ary and binary relations. 


\noindent
\textbf{Dataset.} 
We conduct n-ary relational KBC experiments on three public datasets: 
(1) WikiPeople \citep{guan2019link} is a dataset with over 47k entities and 707 relations in arities of 2-9; 
(2) JF17K \citep{wen2016representation} is a dataset with over 28k entities and 322 relations in arities of 2-6; 
and (3) FB-AUTO \citep{fatemi2019knowledge} is a dataset constructed from Freebase \citep{bollacker2008freebase} with subject of "automotive" and relations in arities of 2, 4, 5. 
The binary relational facts in each dataset are extracted for binary relational KBC experiments. 
We also evaluate RAM on two binary benchmark datasets: (1) WN18 \citep{bordes2013translating} and (2) FB15k \citep{bordes2013translating}.
Since JF17K lacks a valid set, 
we randomly select 20\% of the train set as validation. 
Other datasets follow the split of the corresponding original papers, 
and the detailed statistics are provided in Table~\ref{tab:datasets}.

\noindent
\textbf{Baselines:} As for n-ary relational KBC, 
we compare RAM with the state-of-the-art approaches, 
including the translational model, RAE \citep{zhang2018scalable}, 
the neural network models, NaLP \citep{guan2019link}, HINGE \citep{rossobeyond} and NeuInfer \citep{guan2020neuinfer}, 
and the multilinear model, HypE \citep{fatemi2019knowledge}. 
Although Freebase \citep{bollacker2008freebase} introduces auxiliary nodes  to reify n-ary relational data into binary relational data, 
such nodes are unavailable for evaluation, thus the comparison on reified data is omitted. 
Besides, GETD \citep{liu2020generalizing} can only model single-arity relational KBs, 
while StarE \cite{StarE} focuses on triplet in n-ary relations, 
thus they are not included in comparison. 
For binary relational KBC, we compare RAM with several strongest baselines, including TransE \citep{bordes2013translating}, DistMult \citep{wang2017knowledge}, ComplEx \citep{trouillon2016complex}, SimplE \citep{kazemi2018simple}, RotatE \citep{sun2019rotate} and TuckER \citep{balazevic2019tucker}.
We implemented RAM using PyTorch. 
We report the better results between original paper reported and that obtained by our fine-tune.

\noindent
\textbf{Evaluation Metrics:} 
Two standard metrics are used for evaluation, mean reciprocal rank (MRR) and Hit@$k$ with $k$=1, 3, 10. 
Especially, both metrics are in filtered setting \citep{bordes2013translating}, 
and entities at all positions of the fact are calculated for metrics \citep{fatemi2019knowledge,guan2019link,guan2020neuinfer}.

\noindent
\textbf{Implementation:}
The implementation of RAM is available
at Github\footnote{\url{https://github.com/liuyuaa/RAM}}. 
For simplicity, we take $m=2,\ K=10$ in experiments. 
The batch size is set to 64 and the embedding size $d$ is determined by hardware resources among \{25, 50\}. 
All other hyper-parameters are tuned over the validation set with early stopping used. 
The learning rate is selected from \{0.005, 0.003, 0.002, 0.001\} and the decay rate from \{0.995, 0.99\}. 
Dropout is used for regularization, 
selected from \{0.0, 0.2, 0.4\}. 
All experiments are run on a single Titan-XP GPU.

\subsection{Benchmark Comparison} \label{sec:multiKBC}

\begin{figure*}[ht]
	\centering
	\subfigure[WikiPeople]{ \label{fig:bar_WikiPeople} 
		\includegraphics[width=0.28\textwidth]{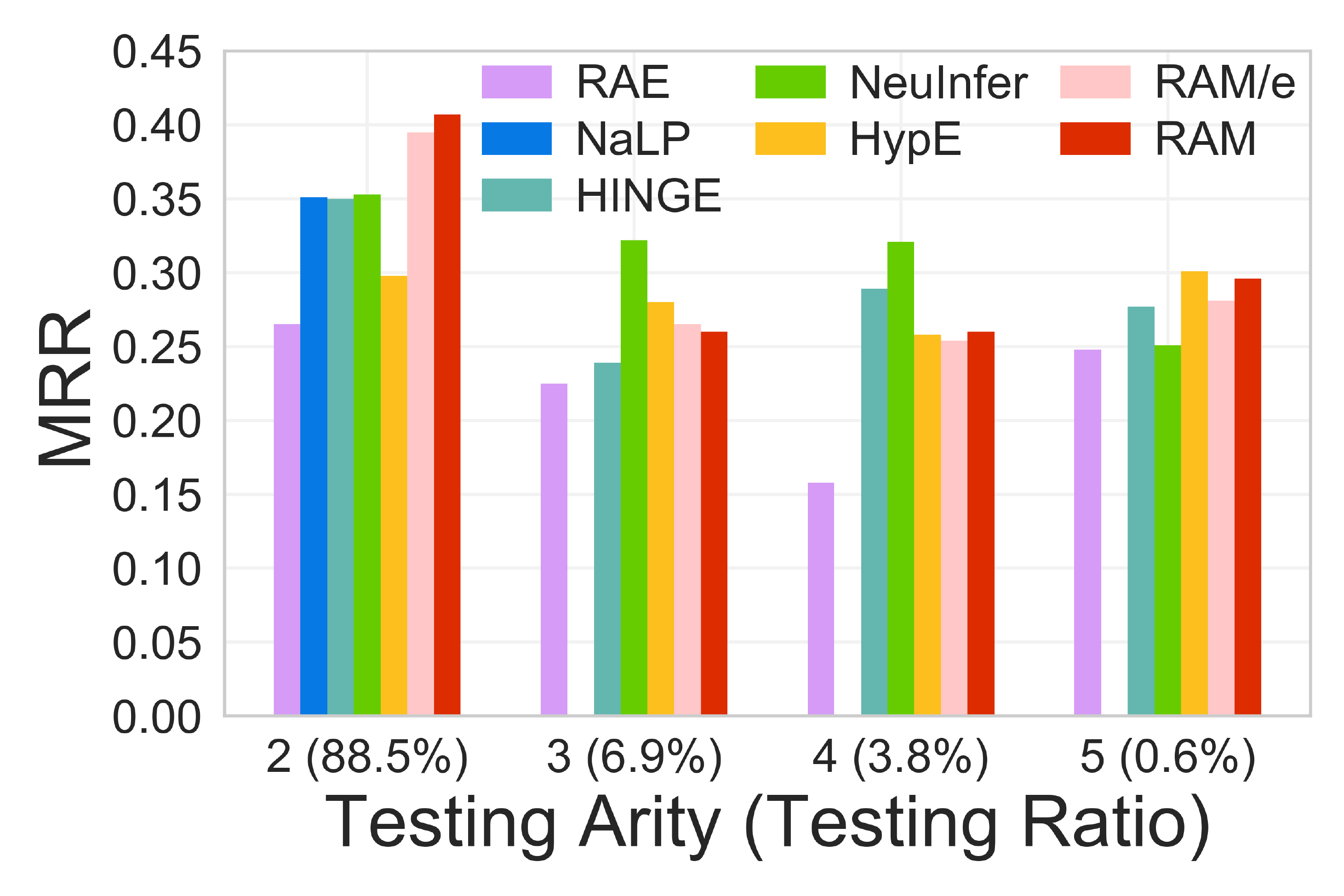}}
	\quad
	\subfigure[JF17K]{ \label{fig:bar_JF17K} 
		\includegraphics[width=0.28\textwidth]{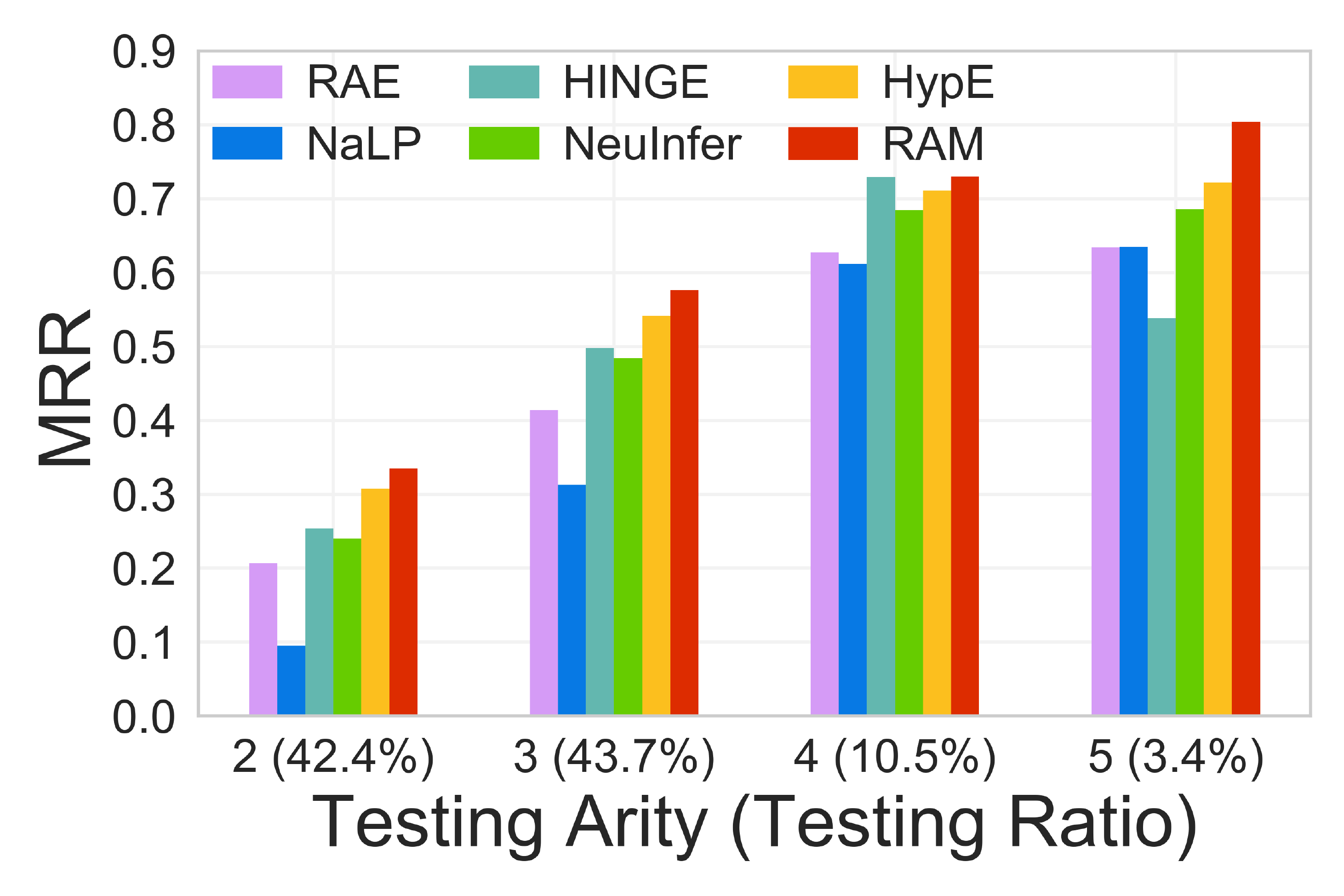}}
	\quad
	\subfigure[FB-AUTO]{ \label{fig:bar_FB-AUTO} 
		\includegraphics[width=0.28\textwidth]{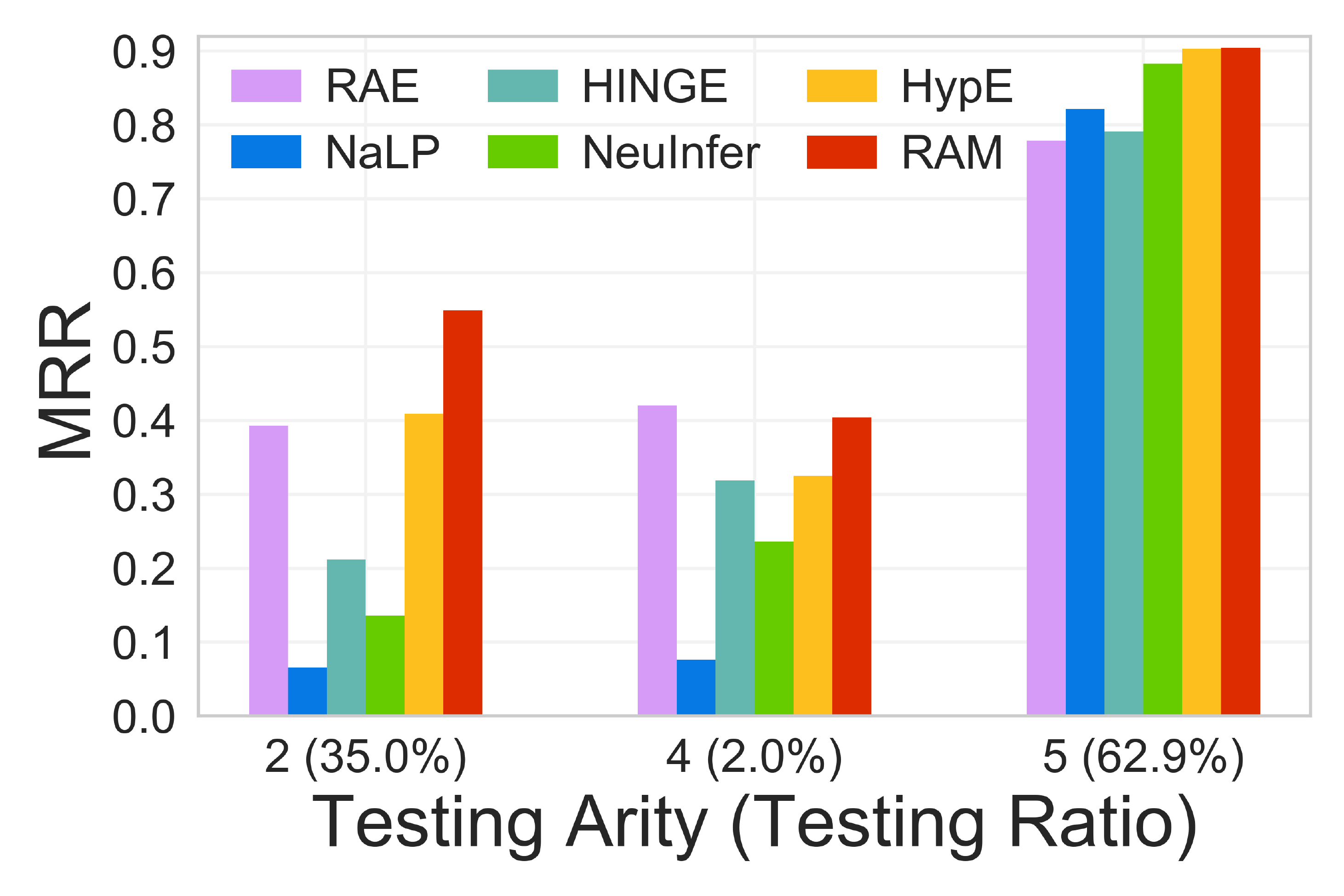}}
	\vspace{-12px}
	\caption{Breakdown performance across relations with different arities. $x$-axis identifies the relation arity and the ratio of testing samples. 6-ary relational facts and beyond are few and unreliable, thus not reported. Beyond-binary results of NaLP on WikiPeople are not provided in \citep{guan2019link}, thus not reported.}
	\label{fig:breakdown}
\end{figure*}

Table~\ref{tab:resuls_n-ary} shows the empirical results on n-ary relational datasets, 
and the breakdown performance across arities are presented in Figure~\ref{fig:breakdown}. 
Since explicit hand-annotated roles are available in WikiPeople, 
we drop the latent space of RAM and assign each role with an embedding vector and a pattern matrix, referred to as RAM/e.

From Table~\ref{tab:resuls_n-ary}, 
we observe that RAM improves state-of-the-art MRR by at least 0.03 on all three datasets. 
On the other hand, NaLP and HINGE are relatively weak due to the complex architectures with overfitting, 
while RAE and HypE are limited by only relation level modeling. 
In particular, on the hardest WikiPeople dataset with the most entities and relations, 
both RAM and RAM/e significantly outperform baselines. 
These results provide evidence for our effective role-aware modeling. 
Moreover, the better performance of RAM than RMA/e implies that the latent space better captures the semantic relatedness than explicit roles in n-ary relational KBs.

As for breakdown performance in Figure~\ref{fig:breakdown}, 
RAM performs consistently well across different arities on JF17K and FB-AUTO, 
while the relatively weak performance on higher-arity WikiPeople data is due to the unbalance data distribution in dataset, 
which is discussed in Appendix~\ref{app:exp}. 
Moreover, RAM obtains the significant results on binary relational data, 
which further validates the generalization power in binary relational KBs.

\begin{figure*}[ht]
	\centering
	\subfigure[WikiPeople]{ \label{fig:time_wikipeople} 
		\includegraphics[width=0.28\textwidth]{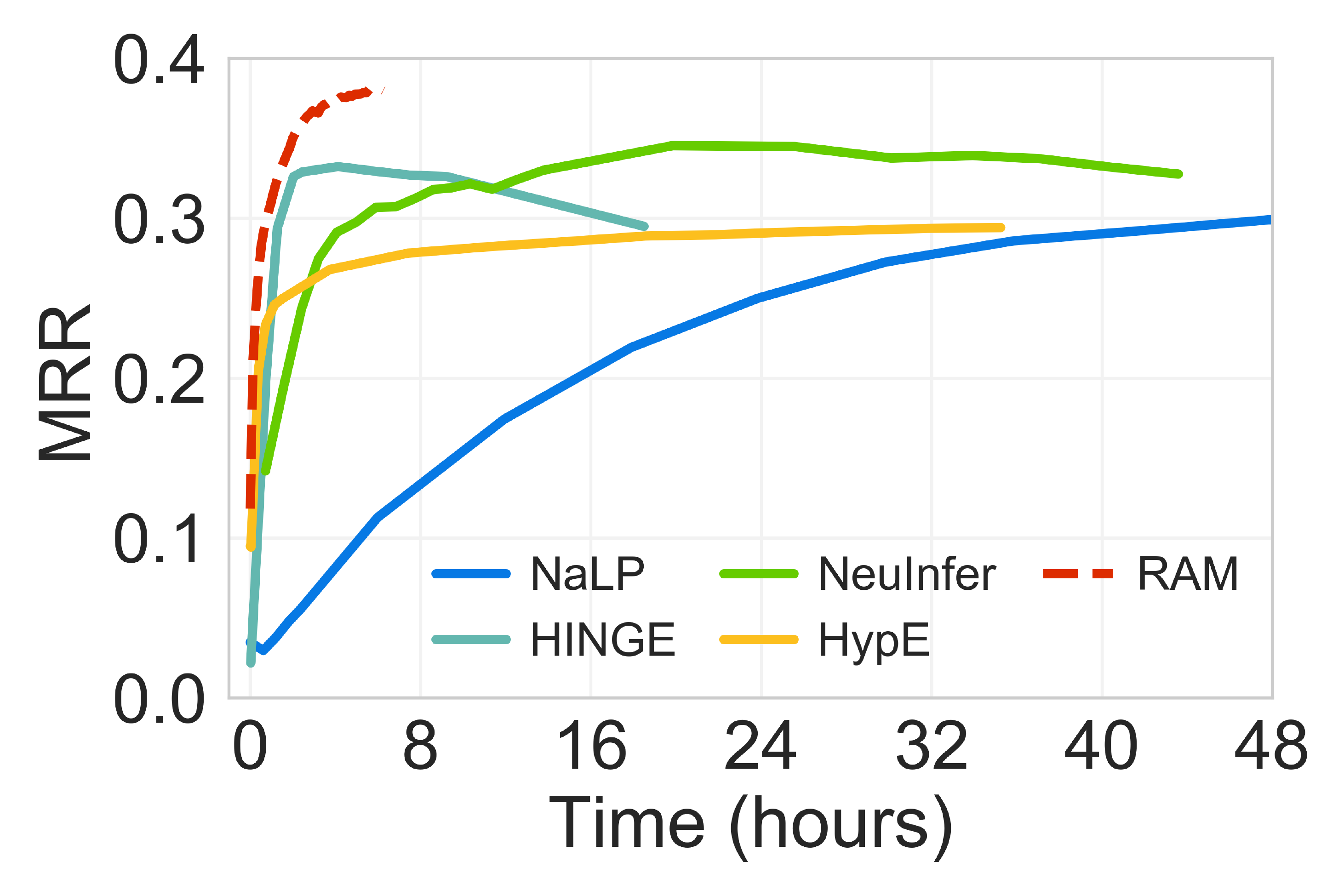}}
	\quad
	\subfigure[JF17K]{ \label{fig:time_jf17k} 
		\includegraphics[width=0.28\textwidth]{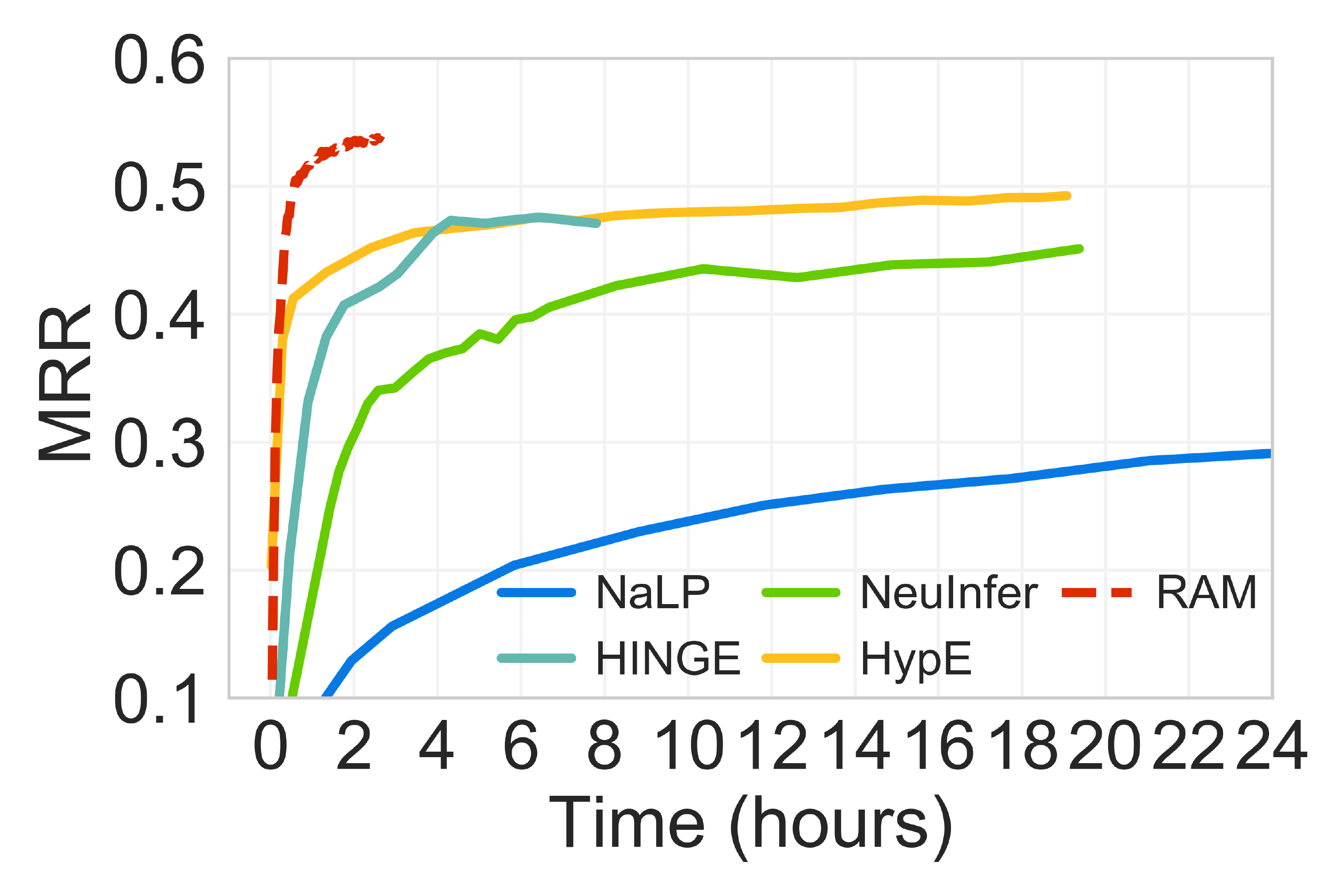}}
	\quad
	\subfigure[FB-AUTO]{ \label{fig:time_fbauto} 
		\includegraphics[width=0.28\textwidth]{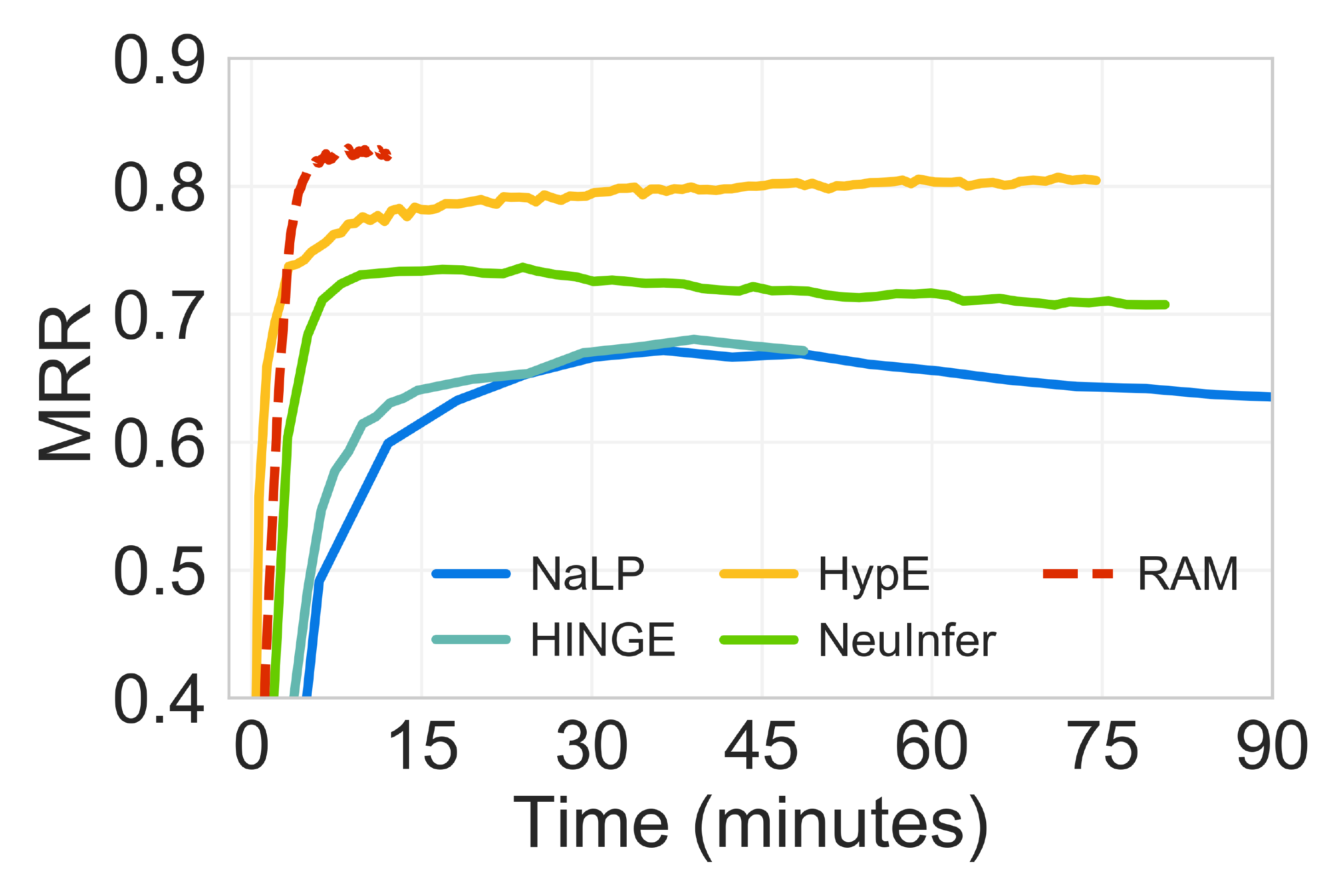}}
	\vspace{-12px}
	\caption{Comparison on clock time of model training vs. testing MRR between RAM and baselines.}
	\label{fig:time}
\end{figure*}

Besides, we plot the learning curves of baselines and RAM on three datasets in Figure~\ref{fig:time}. 
With straightforward multilinear product form, 
RAM outperforms all the baselines and converges much faster, 
which is consistent with the linear time complexity shown in Table~\ref{tab:MultiAry_SF}. 
Note that due to the time-consuming pairwise relatedness calculation, 
the training of NaLP is much slower than other models on large datasets WikiPeople and JF17K.

\begin{table*}[ht]
	\setlength{\tabcolsep}{2pt}
	\caption{Binary relational KBC results. Best results are in bold and second best results are underlined. For WN18 and FB15k, results of TransE are copied from \citep{sun2019rotate} and results of DistMult are copied from \citep{balazevic2019tucker}, and other results are copied from the corresponding original papers. Note that RAM/b and RAM are same on WN18 and FB15k.} \label{tab:results_binary_1}
	\centering
	\label{tab:mbkbin}
	\vspace{-10px}	
	\begin{tabular}{cccc|ccc|ccc|ccc|ccc}
		\toprule		
		& 
		\multicolumn{3}{c}{\textbf{WikiPeople}} 
		& \multicolumn{3}{c}{\textbf{JF17K}}                                                                                                                         & \multicolumn{3}{c}{\textbf{FB-AUTO}}       
		& \multicolumn{3}{c}{\textbf{WN18}}                                                                                                                         & \multicolumn{3}{c}{\textbf{FB15k}}                                                                                                                  \\
		\textbf{Model} &{MRR}& {Hit@10} & {Hit@1} &{MRR}& {Hit@10} & {Hit@1} & {MRR} & {Hit@10}  & {Hit@1} & {MRR} & {Hit@10}  & {Hit@1} & {MRR} & {Hit@10}  & {Hit@1} \\
		\midrule
		TransE \citep{bordes2013translating}
		& 0.312 &\textbf{0.574} & 0.146 
		& 0.276 & 0.495 & 0.167 
		& 0.313 & 0.562 & 0.132 
		& {0.495} & {0.943} & {0.113} 
		& {0.463} & {0.749} & {0.297}  \\
		DistMult \citep{wang2017knowledge}
		& 0.275 & 0.388 & 0.193 
		& 0.228 & 0.411 & 0.144 
		& 0.494 & 0.566 & 0.444
		& {0.822} & {0.936}  & {0.728}
		& {0.654} & {0.824}   & {0.546}     \\
		ComplEx \citep{trouillon2016complex} 
		& 0.326 & 0.461 & 0.232 
		& 0.308 & 0.498 & 0.219 
		& 0.487 & 0.568 & 0.442
		& {0.941}          & {0.947}       & {0.936}    
		& {0.692}          & {0.840}       & {0.599}     \\
		SimplE \citep{kazemi2018simple}
		& 0.326 & 0.449 & 0.249 
		& 0.313 & 0.502 & 0.224 
		& 0.493 & 0.577 & 0.440
		& {0.942}          & {0.947}               & {0.939}    
		& {0.727}          & {0.838}                 & {0.660}   \\
		RotatE \citep{sun2019rotate} 
		& 0.422 & 0.519 &	0.285 
		& 0.304 & 0.496 & 0.210 
		& 0.470&	0.577& 0.408
		& \underline{0.949}          & {\textbf{0.959}}   & \underline{0.944} 
		& {\underline{0.797}}   & \underline{0.884}  & \underline{0.746}		 \\
		TuckER \citep{balazevic2019tucker}
		& \underline{0.429}  & 0.538 &\underline{0.365} 
		& \underline{0.333}&\underline{0.512}	&\underline{0.244}
		&	0.510&\underline{0.621}	&	0.450
		& {\textbf{0.953}} & {\underline{0.958}}  & {\textbf{0.949}} 
		& {0.795}          & {\textbf{0.892}}   & {0.741} \\
		\midrule
		{RAM/b} 
		& \textbf{0.445} & {0.562} &\textbf{0.374}
		& 0.324 & 0.508& 0.234 & \underline{0.518}
		& 0.604 & \underline{0.468}  & \multirow{2}{*}{0.947} & \multirow{2}{*}{0.952} & \multirow{2}{*}{0.943} & \multirow{2}{*}{\textbf{0.803}} & \multirow{2}{*}{0.882} & \multirow{2}{*}{\textbf{0.756}}   \\
		{RAM}   
		&0.408  & \underline{0.568} & 0.303 
		& {\textbf{0.337}} & {\textbf{0.523}}  & {\textbf{0.246}}
		& {\textbf{0.557}} & {\textbf{0.649}}  & \textbf{0.507}
		&    &  &  & &   & 	   \\
		\bottomrule
	\end{tabular}
\end{table*}

\subsection{The Impact of Mixed-arity Relational Data to Single-arity Relational KBC}\label{sec:single vs nary}
The above results are evaluated on n-ary relational KBC, 
in respect of mixed-arity relational data. 
Thus, an obvious question is, 
if the mixed n-ary relational data can improve the modeling of single-arity relational KBs, 
like binary relational KBs \citep{rossi2020knowledge,wang2017knowledge} and higher-arity relational KBs in GETD \citep{liu2020generalizing}. 
We investigate both cases here.

\subsubsection{The Impact to Binary Relational KBC}
\label{sec:bin}
For the binary relational KBC task, 
we extract binary relational data from n-ary relational datasets (WikiPeople, JF17K, FB-AUTO). 
Note that RAM is trained on n-ary relational data and evaluated on extracted binary relational data, 
while other baselines are consistently trained and evaluated on extracted binary relational data. 
RAM/b is the version of RAM trained only on binary relational data.

According to the results in Table~\ref{tab:results_binary_1}\footnote{Due to the space limitation, results with Hit@3 are omitted here, which are in accord with other metrics. }, 
RAM outperforms baselines on JF17K and FB-AUTO, 
while RAM/b achieves the best results on WikiPeople, 
which indicate the benefits of the proposed role-aware modeling. 
Moreover, 
the superior performance of RAM demonstrates that n-ary relational data improve the modeling of binary relational KBs, 
by sharing information across different arities. 
For the performance gap between RAM/b and RAM on WikiPeople, 
we discuss the influence of binary relational data ratio in Appendix~\ref{app:exp}.

To further validate the effectiveness as well as robustness, 
the results on binary benchmark datasets, WN18 and FB15k, 
are also shown in Table~\ref{tab:results_binary_1}. 
According to the results, 
RAM achieves close performance to the best approaches on WN18. 
Moreover, RAM achieves the best results on FB15k, the hard dataset with a very large number of relations. 
Such results further validate Proposition~\ref{theorem:conblm} that RAM can generalize several bilinear models in binary relational KBs with comparable performance.
We emphasize that, 
most binary relational KBC approaches cannot handle beyond-binary relations, 
while n-ary relational approaches like HypE are still uncompetitive in binary relational KBC. 
Therefore, RAM provides effective generalized modeling for both binary and n-ary relational KBs.

\subsubsection{The Impact to Higher-arity Relational KBC}\label{sec:single-ary}
Since the model GETD \citep{liu2020generalizing} considers single-arity relational KBs with higher-arity relations, 
in Table~\ref{tab:results_multi4Nary}, 
we investigate if mixed-arity relational data improves the modeling of higher-arity relational KBs.
Specifically, 
in the first row, 
each value represents the result of a GETD model trained and evaluated on corresponding arity of relational data. 
For example, a GETD model is learned over 5-ary relational data of FB-AUTO, 
which achieves the MRR of 0.786. 
On the other hand, in the second row, 
a RAM model is trained on n-ary relational data, 
and results of each arity are reported. 
It can be observed that RAM performs better on most single-arity relational KBs especially the high arity ones, 
which again validates the sharing information across roles. 
The slightly weaker result on 4-ary relational data of JF17K is mainly affected by noise introduced by other arities of relational data. 
Such results also indicate the importance of research in n-ary relational KB modeling with mixed arities considered.

\begin{table}[htbp]
	\caption{Higher-arity relational KBC results on JF17K and FB-AUTO, where best results are in bold. The binary case is also included for comparison.}\label{tab:results_multi4Nary}
	\centering
	\vspace{-10px}
	\setlength\tabcolsep{2pt}
	\begin{tabular}{cccccc|ccc}
		\toprule
		& \multicolumn{5}{c|}{\textbf{JF17K}} & \multicolumn{3}{c}{\textbf{FB-AUTO}} \\
		\textbf{Model}  & 2    & 3    & 4   & 5      & 6   & 2  & 4   & 5    \\
		\midrule
		{GETD\citep{liu2020generalizing}} & \textbf{0.339}          & { \textbf{0.583}} & {\textbf{0.751}} & {{0.746}} & {0.350}          &{ 0.524}          & {0.237}          & {0.786}          \\
		\textbf{RAM}   & {0.337} & {0.578} & { 0.736}          & \textbf{0.805}          & {\textbf{0.697}} &  \textbf{0.557} & \textbf{0.456} & {\textbf{0.904}}\\
		\bottomrule
	\end{tabular}
\end{table}

\begin{figure}[ht]
	\centering
	\includegraphics[width=0.82\linewidth]{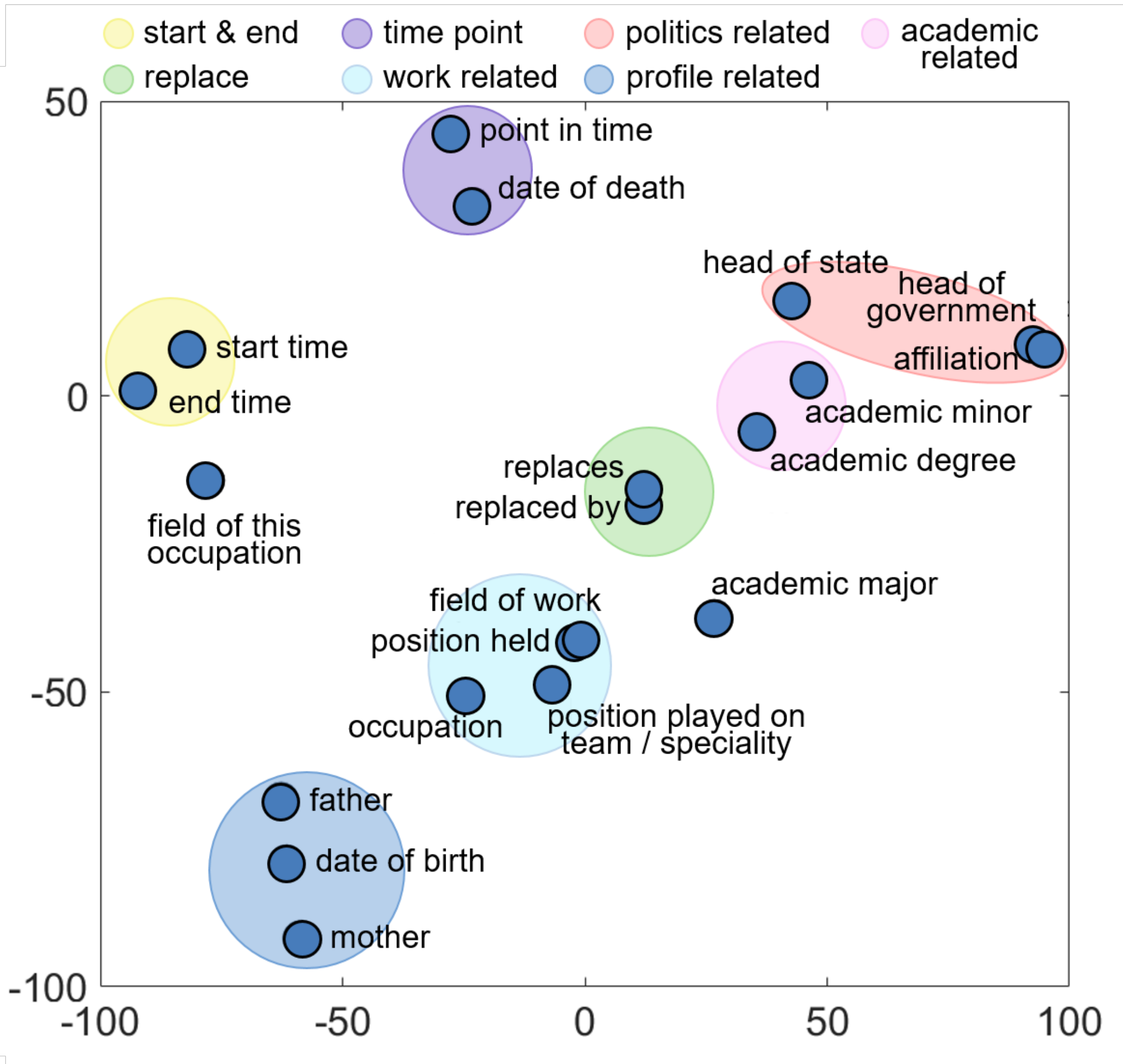}
	\vspace{-12px}
	\captionof{figure}{t-SNE of learned role embeddings on WikiPeople.}
	\label{fig:role_vis}
\end{figure}

\begin{figure*}[ht]
	\centering
	\subfigure[FB15k/spouse]{ \label{fig:fb15k-spouse} 
		\includegraphics[width=0.25\textwidth]{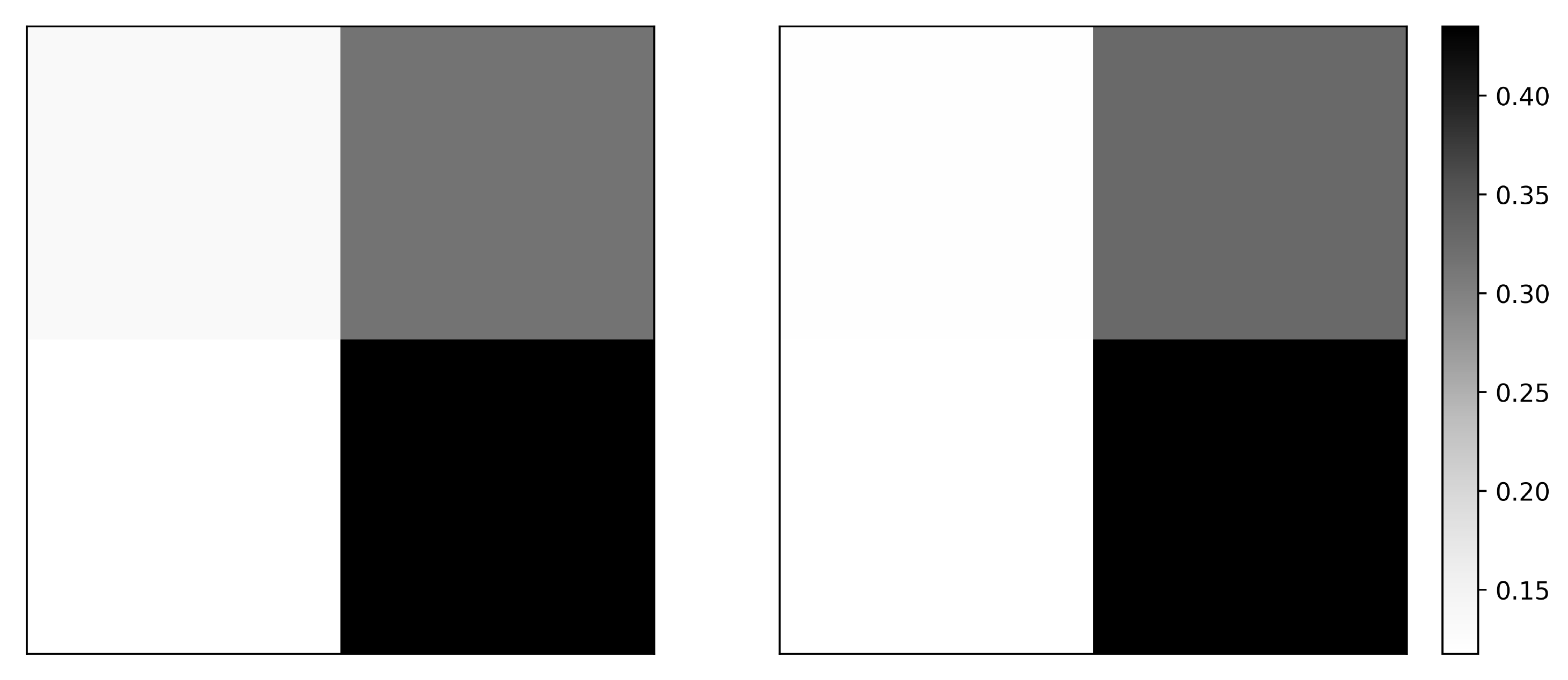}}
	\qquad
	\subfigure[WN18/hypernym]{ \label{fig:wn18-hyper} 
		\includegraphics[width=0.25\textwidth]{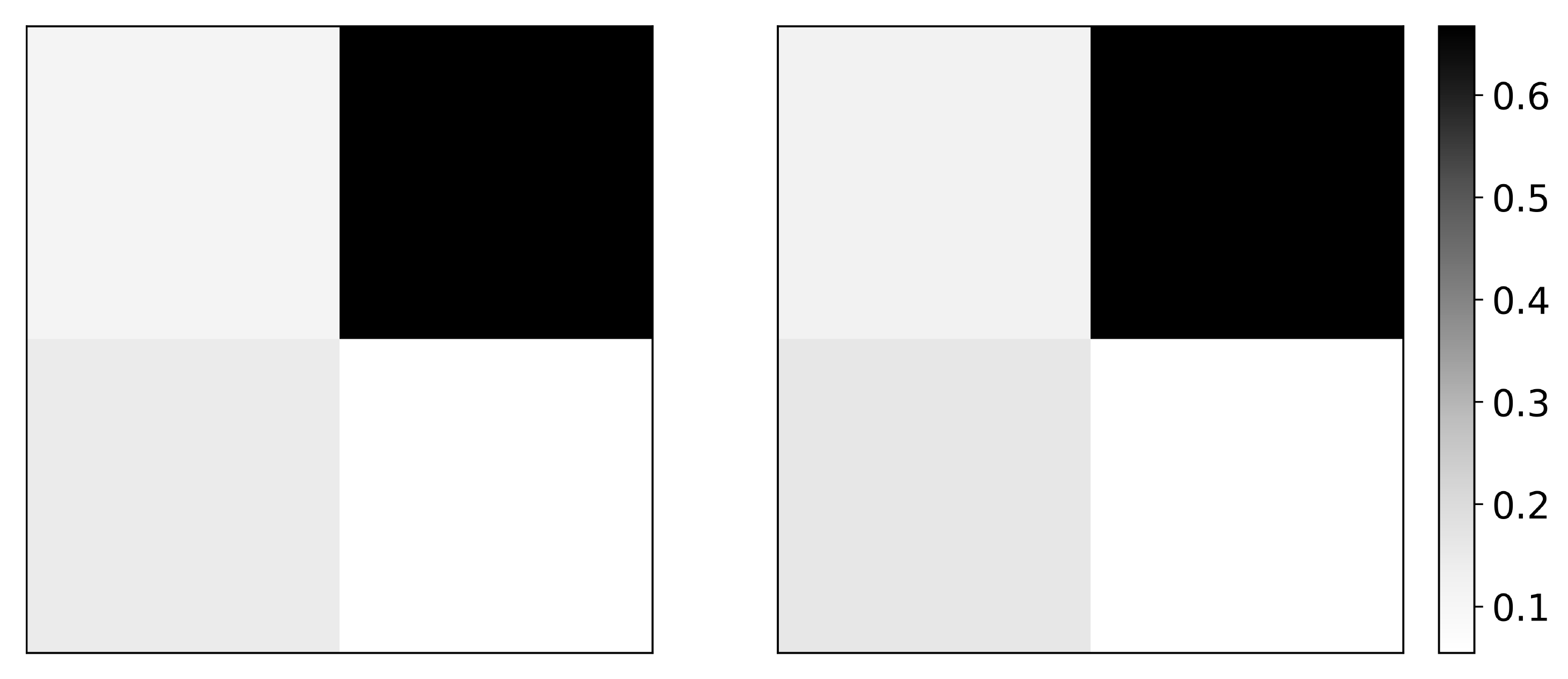}}
	\qquad
	\subfigure[WN18/hyponym]{ \label{fig:wn18-hypo} 
		\includegraphics[width=0.25\textwidth]{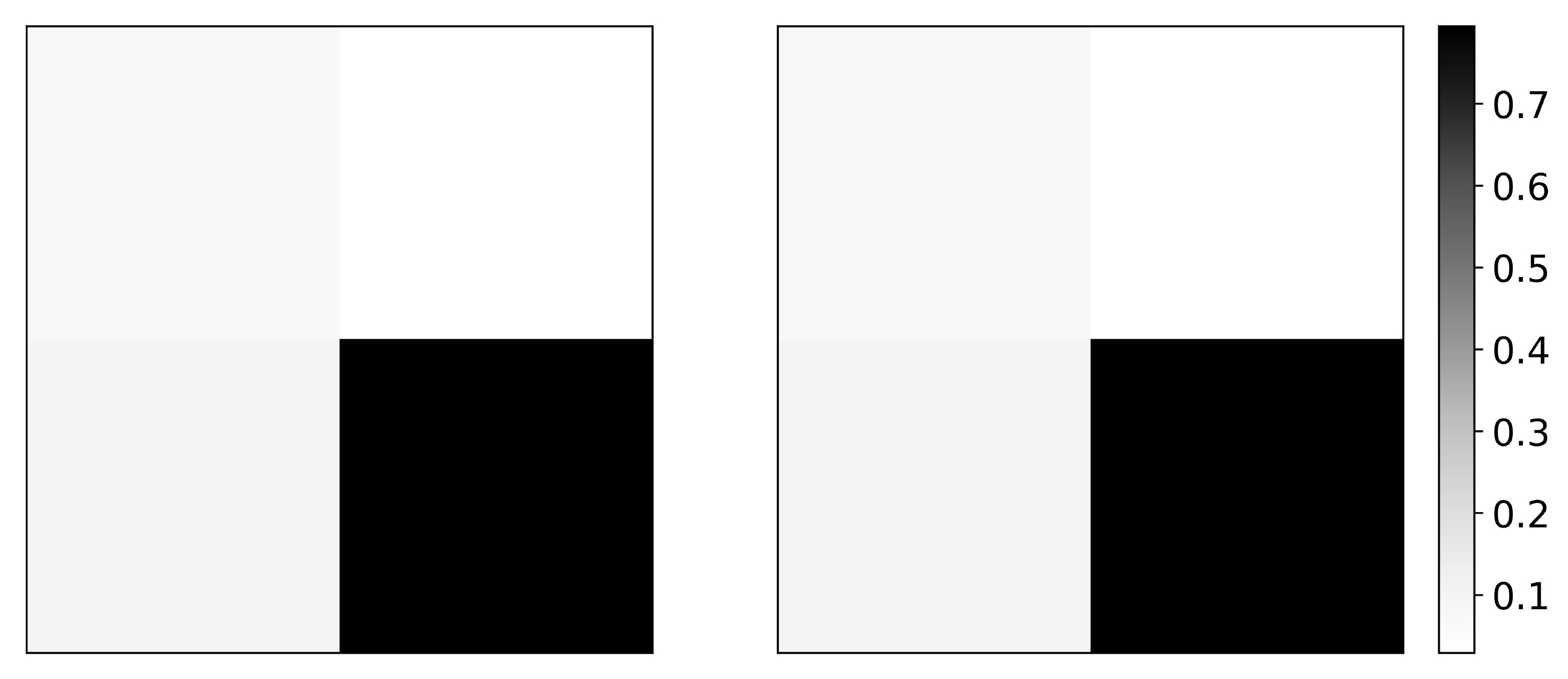}}
	\vspace{-12px}
	\caption{Role-aware pattern matrix visualization of binary relations in FB15k and WN18, with symmetric and inverse patterns.}
	\label{fig:binary-symmetric_inverse_relations}
\end{figure*}

\begin{figure*}[ht]
	\centering
	\subfigure[MRR v.s. $d$]{ \label{fig:mrr_d} 
		\includegraphics[width=0.24\textwidth]{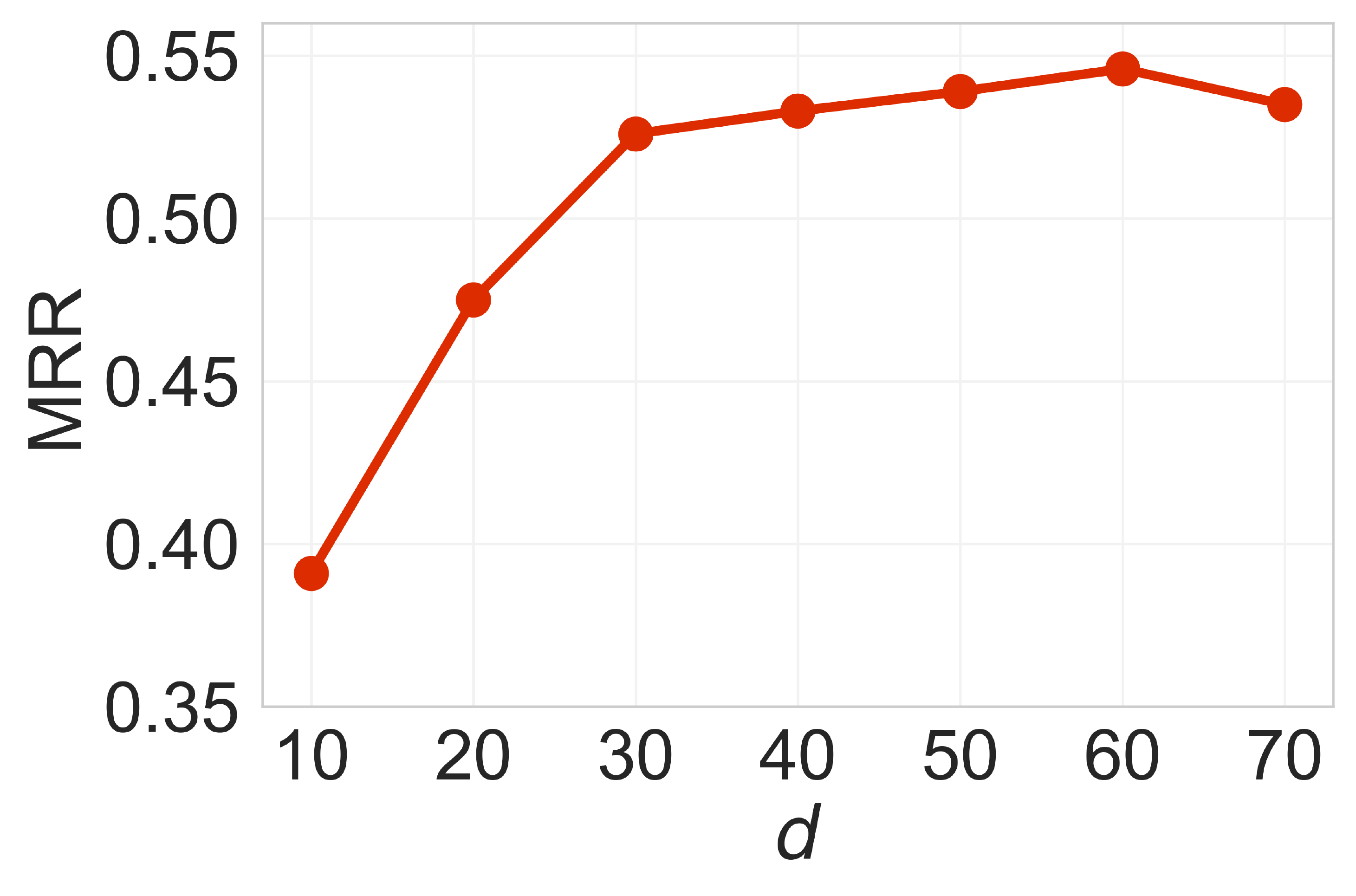}}
	\qquad
	\subfigure[MRR v.s. $K$]{ \label{fig:mrr_K} 
		\includegraphics[width=0.25\textwidth]{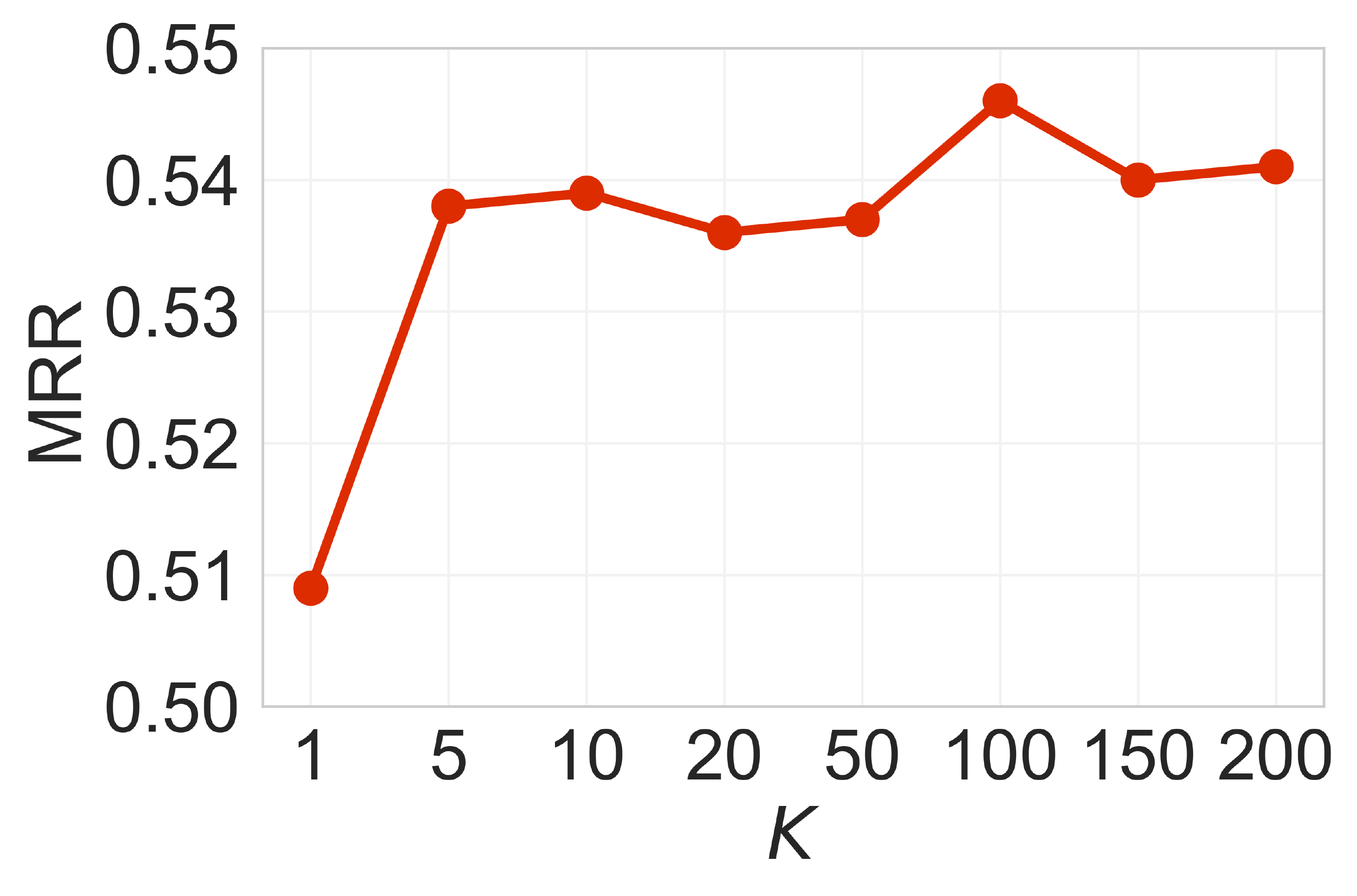}}
	\qquad
	\subfigure[MRR v.s. $m$]{ \label{fig:mrr_m} 
		\includegraphics[width=0.25\textwidth]{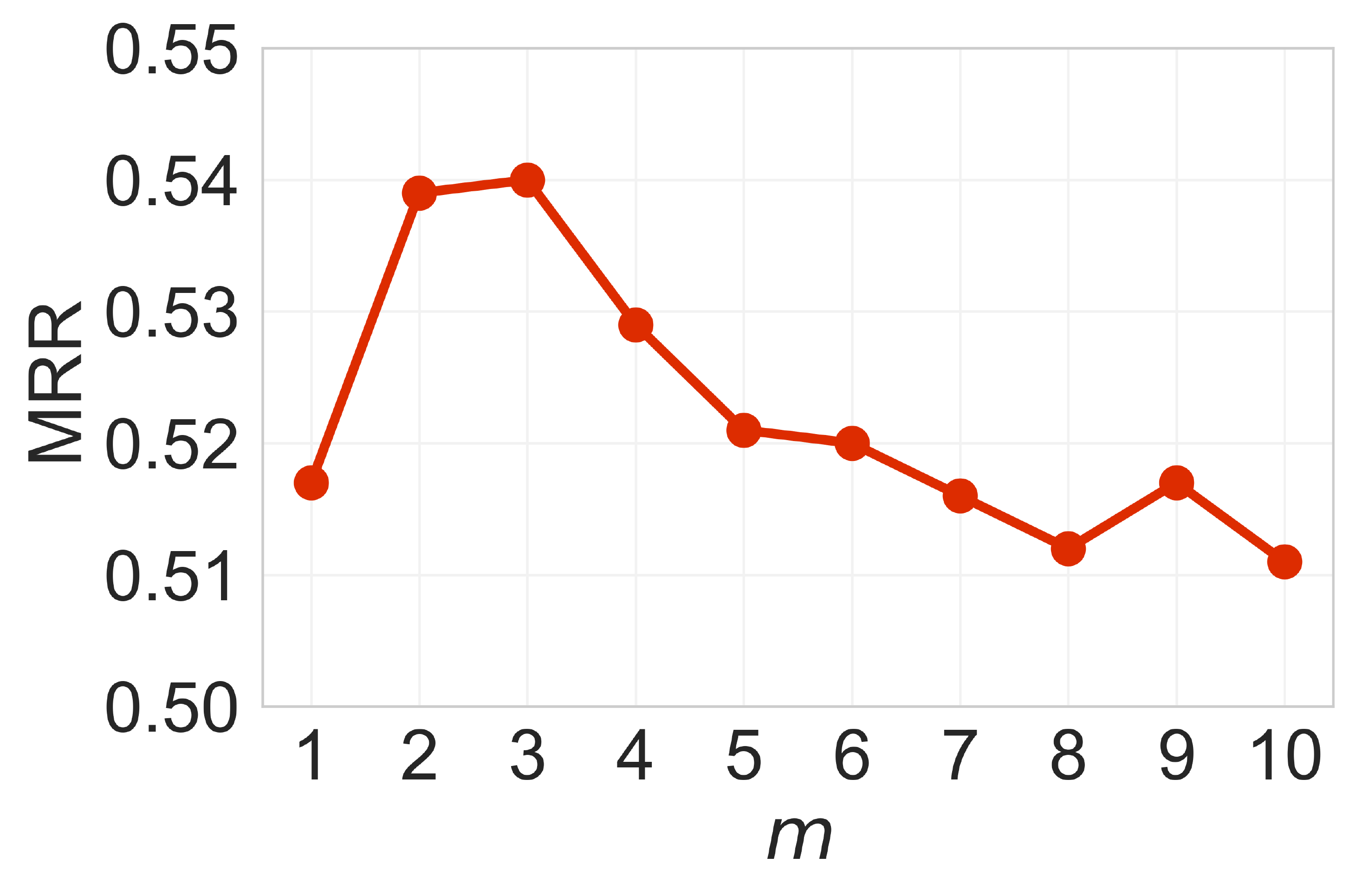}}
	\vspace{-12px}
	{\caption{Effects of (a) embedding dimensionality $d$, (b) latent space size $K$, and (c) multiplicity of entity embeddings $m$ on the testing MRR on JF17K dataset.}
		\label{fig:ablation_study}}
\end{figure*}

\subsection{Role-Aware Visualization}\label{sec:vis}

To validate the role-aware modeling of RAM, 
we visualize the learned role embeddings on WikiPeople dataset using t-SNE \citep{maaten2008visualizing}. 
We manually select 20 semantically related roles, and show in Figure~\ref{fig:role_vis}. 
We observe that most semantically related roles are clustered in space, 
such as \emph{academic minor} and \emph{academic degree}, 
\emph{start time} and \emph{end time}, etc. 
Since the embedding size on WikiPeople is only 25, 
there may be inaccurate points like \emph{academic major} in Figure~\ref{fig:role_vis}.

Furthermore, we also investigate the learned role-aware pattern matrix in RAM. 
Since relations are represented in the role level, 
roles also inherit the typical relation patterns such as symmetric and inverse patterns in binary relations. 
Specifically, the symmetric pattern means that the role has the same compatibility to all involved entities, 
i.e., each row of the pattern matrix should be identical. 
As for the inverse pattern, 
the compatibility between the role and two entities should be exchanged for its inverse role, 
i.e., exchanging rows of the pattern matrix for one role leads to the pattern matrix for its inverse role. 
Especially, such pattern capture is not available in existing approaches, 
which only consider the compatibility between the role and its mapping entity.

We evaluate RAM on WN18 and FB15k with typical symmetric and inverse patterns. 
Figure~\ref{fig:fb15k-spouse} visualizes the pattern matrices of two roles in symmetric relation \texttt{spouse} in FB15k, 
where both rows of pattern matrices are the same. 
Note that pattern matrices for different roles can be different for diverse semantics. 
Figure~\ref{fig:wn18-hyper} and (c) show the role-aware pattern matrices of inverse relations \texttt{hypernym} and \texttt{hyponym} in WN18 respectively. 
Especially, the rows of pattern matrices in \texttt{hypernym} are the reverse rows of corresponding pattern matrices in \texttt{hyponym}. 
Therefore, RAM successfully captures typical patterns with strong expressiveness.

\subsection{Sensitivity of Hyper-parameters}\label{sec:ablation}

Figure~\ref{fig:ablation_study} further investigates the influence of key hyper-parameters on JF17K dataset, 
including embedding dimensionality $d$, latent space size $K$, and multiplicity of entity embeddings $m$.
Since the full expressiveness guarantees the learning capacity, 
we can observe that RAM achieves consistent good performance when $d$ is over 30 in Figure~\ref{fig:mrr_d}. 
As for Figure~\ref{fig:mrr_K}, a small latent space size ($K\geq5$) is enough for robust performance, 
which validates the effectiveness of roles' semantic relatedness in n-ary relational KBs. 
Besides, there is a peak point in Figure~\ref{fig:mrr_m}. 
Compared with single embedding, the multiplicity of entity embeddings with $m=2$ or $3$ provides better coverage for entity semantics, 
while lager $m$ with more parameters leads to overfitting and intractable learning. 
Based on Proposition~\ref{theorem:conblm}, existing bilinear models also choose small $m$, 
such as $m=2$ in DistMult \citep{wang2017knowledge} and SimplE \citep{kazemi2018simple}. 
Thus, the appropriate value for $m$ is $2\sim3$, while embedding dimensionality and latent space size are mainly determined by the dataset scale.

\section{Conclusion}
In this paper, we proposed a role-aware model RAM for n-ary relational KBs, which learns the knowledge representation from the role level. By leveraging the latent space for roles' semantic relatedness and pattern matrix for role-entity compatibility, it achieves efficient computation, full expressiveness as well as good generalization in n-ary relational KB modeling. Experiment results on both n-ary and binary relational KB datasets demonstrate the superiority and robustness of RAM.

Compared with studies in traditional binary relational KBs, 
n-ary relational KB study is still in its infancy. 
To better model n-ary relational KBs, 
as future work, 
we plan to explore logic reasoning \citep{ren2019query2box} and rule mining \citep{sadeghian2019drum}, 
which have been shown important in binary relational KBs. 
Moreover, we also consider the typical application of question answering over n-ary relational KBs.


\begin{acks}
This work was supported in part by The National Key Research and Development Program of China under grant 2020AAA0106000, the National Natural Science Foundation of China under U1936217, 61971267, 61972223, 61941117, 61861136003
\end{acks}


\appendix

\section{Role in Other Research Domains}\label{app:role_work}
In the area of graph mining \cite{rossi2014role}, 
the word \emph{role} is also widely used, however, 
they are conceptually different from our role-aware modeling. 
Specifically, the concept is mainly related to role discovery in networks \cite{rossi2014role}, 
which focuses on mining graph nodes with similar connectivity patterns, 
e.g., star-center/edge, peripheral, near-clique nodes, and motif/graphlet \cite{pei2019infinite}. 
Besides, \cite{ahmed2017edge} tries to discover edge role via motifs. 
Overall, these works mainly focus on topology structures of graphs, 
where the role is a structure concept. 
However, in our role-aware modeling, 
roles are semantic compositions of relations, 
where the role is a semantic concept. 

On the other hand, 
the semantic concept of roles in other research domains could motivate our role-aware modeling. 
For example, 
both the influential linguistic project FrameNet \citep{baker1998berkeley} and standard NLP task of semantic role labeling \citep{gildea2002automatic} stress the importance of semantic roles to their applications. 
In the computer vision area, 
a recent work \cite{zareian2020weakly} also adopts the concept of semantic roles similar to ours for visual semantic parsing tasks.

\section{Theoretical Details}

\subsection{Proof of Theorem 1} \label{app:full expressive}
\begin{proof}
	Let $\mathcal{F}^{\text{all}}$ be the set of all true facts in the n-ary relational KB with $\gamma=\vert\mathcal{F}^{\text{all}}\vert$. Then, the statement of Theorem~\ref{theorem:full expressive} is equivalent to assign the parameters 
	(entity embeddings $\{\bm{E}_i\}$, role basis vectors $\{\hat{\bm{u}}_{i}\}$, pattern basis matrices $\{\hat{\bm{P}}_{i}\}$ and role weights $\{\bm{\alpha}^{r}_{i}\}$) 
	of RAM such that the scoring function can be expressed as,
	\begin{align}
		\!\!\!
		\phi(x)
		\begin{cases}
			>0&\mbox{if  $x\in\mathcal{F}^{\text{all}}$}\\
			=0&\mbox{if  $x\notin\mathcal{F}^{\text{all}}$}
		\end{cases},
		\ \text{for} \ x\coloneqq\{\gamma^{r}_{1}:e_1,\cdots,\gamma^{r}_{{a_r}}:e_{a_r}\},\label{eq:express-form}
	\end{align}
	under the conditions of embedding dimensionality $d=\eta$, multiplicity of entity embeddings $m=\max_{r\in\mathcal{R}}a_r$, and latent space size $K=\eta$.
	
	We consider the case that $\eta\geq1$, 
	and describe an optional assignment for parameters: 
	For each entity $e\in\mathcal{E}$ with multiple embeddings $\bm{E}\in\mathcal{R}^{m\times d}$, 
	$\bm{E}[i,j]$ is set to 1 if the entity $e$ involves with the $i$-th role of the $j$-th fact in $\mathcal{F}^{\text{all}}$, and to 0 otherwise. 
	As for the latent space, 
	the role basis vectors are concatenated as an identity matrix, 
	i.e., $[\hat{\bm{u}}_1,\cdots,\hat{\bm{u}}_K]=\bm{I}_K$. 
	All the pattern basis matrices are set in the form of $\hat{\bm{P}}_i=[\bm{I}_{a_r},\bm{0}]\in\{0,1\}^{a_r\times m}$. 
	Since $[\hat{\bm{u}}_1,\cdots,\hat{\bm{u}}_K]$ is a group of basis vectors for $\bm{R}^\eta$, 
	the role weights $\{\bm{\alpha}^{r}_{i}\}$ can be assigned to satisfy that $\bm{u}^r_i[j]=1$ if the relation $r$ involves with the $j$-th fact in $\mathcal{F}^{\text{all}}$, and $\bm{u}^r_i[j]=0$ otherwise.
	
	Following the assignment above, 
	for the $j$-th true fact $x\coloneqq\{\gamma^{r}_{1}:e_1,\cdots,\gamma^{r}_{{a_r}}:e_{a_r}\}$, 
	RAM calculates the score as:
	\begin{align}
		\phi(x)= \sum\nolimits_{i=1}^{a_r}  \langle \bm{u}^{r}_{i}, {{\bm{P}}^{r}_{i}{[1,:]}}\bm{E}_1, \cdots, {\bm{P}}^{r}_{i}[a_r,:]\bm{E}_{a_r} \rangle,
	\end{align}
	where $\bm{u}^r_i{[j]}$ is equal to one, and the $j$-th elements of ${\bm{P}}^{r}_{i}{[1,:]}{\bm{E}}_1,\cdots,$ ${\bm{P}}^{r}_{i}{[a_r,:]}\bm{E}_{a_r}$ are all equal to one. 
	Each summation term is equal to one and the final score for $x$ is $\phi(x)=a_r>0$.
	
	To show $\phi(x)=0$ when $x\notin\mathcal{F}^{\text{all}}$, 
	we prove by contradiction. 
	Assume that there exists a false fact $x\coloneqq\{\gamma^{r}_{1}:e_1,\cdots,\gamma^{r}_{{a_r}}:e_{a_r}\}\notin\mathcal{F}^{\text{all}}$ such that $\phi(x)>0$ (Note that $\phi(x)$ is non-negative based on the assignment). 
	Based on the assumption, there was at least a position $j$ to ensure that 
	$\bm{u}^r_i{[j]}=1$, and the $j$-th elements of ${\bm{P}}^{r}_{i}{[1,:]}{\bm{E}}_1,\cdots,{\bm{P}}^{r}_{i}{[a_r,:]}\bm{E}_{a_r}$ are all equal to one. However, this can only happen when entities $e_1,\cdots,e_{a_r}$ and relation $r$ appear in the $j$-th fact (with $e_i$ linked with $\gamma^r_i$) of $\mathcal{F}^{\text{all}}$, then $x\in\mathcal{F}^{\text{all}}$, contradicting the initial assumption. Thus, if $x\notin \mathcal{F}^{\text{all}}$, RAM obtains $\phi(x)=0$.
	
\end{proof}

\subsection{Illustration of Proposition 1} \label{app:gblm}
Recall that each term in \eqref{eq:score} of scoring function corresponds to the role, 
the multi-embedding mechanism can also be applied in role embedding. 
Moreover, each role can be learned with multiple pattern matrices for compatibility capture as well as stronger expressiveness. 
Specifically, given an n-ary relation $r$ with $a_r$ roles, 
each role $\gamma^r_i$ is mapped to $m_{\gamma}$ embedding vectors, denoted by $\bm{u}^r_{ij}\in\mathbb{R}^d, \forall j\in\{1,\cdots,m_{\gamma}\}$. 
Besides, each role embedding vector $\bm{u}^r_{ij}$ corresponds to $n_{\text{Pmat}}$ pattern matrices, denoted by $\bm{P}^{r}_{ijk}\in\mathbb{R}^{a_r\times m}, \forall k\in\{1,\cdots,n_{\text{Pmat}}\}$. 
Thus, the scoring function of RAM is modified as,
\begin{align*}
&	\phi(x)= \! \\
&	\sum\nolimits_{i=1}^{a_r}
	\sum\nolimits_{j=1}^{m_{\gamma}} 
	\sum\nolimits_{k=1}^{n_{\text{Pmat}}}  
	\!\!\omega_{ijk} \cdot \left\langle 
	\bm{u}^{r}_{ij}, {{\bm{P}}^{r}_{ijk}{[1,:]}}\bm{E}_1, \cdots, {\bm{P}}^{r}_{ijk}[a_r,:]\bm{E}_{a_r} 
	\right\rangle, \notag
\end{align*}
where $\omega_{ijk}\in\mathbb{R}$ is a learnable weight for the role-aware interaction term. 
Note that the originally proposed scoring function \eqref{eq:score} is a special case of above scoring function with $m_{\gamma}=1$, $n_{\text{Pmat}}=1$, and all learnable weights  $\omega_{ijk}$ set to 1.

Now we illustrate the generalization from RAM's modified scoring function to ComplEx and QuatE. 
For ComplEx, each entity is assigned with two embeddings, 
e.g., $\bm{E}_h=[\bm{e}_{h,1};\bm{e}_{h,2}]$, 
and the two roles of binary relation $r$, 
$\gamma^r_1$ and $\gamma^r_2$ correspond to two role embeddings of $\bm{u}^r_{1,1}$ and $\bm{u}^r_{2,1}$ ($m_{\gamma}=1$). 
Especially, each role is assigned with two pattern matrices ($n_{\text{Pmat}}=2$), 
e.g. ${\bm{P}}^{r,\text{CE}}_{1,1,1}$ and ${\bm{P}}^{r,\text{CE}}_{1,1,2}$ for the first role $\gamma^r_1$. 
The weight for each role's relatedness term is set to 1 or -1.
According to \citep{trouillon2016complex}, the scoring function of ComplEx can be rewritten as:
	\begin{align*}
		&\phi(r,h,t)=\frac{1}{4}\text{Re}(\langle \bm{r}, \bm{h}, \bm{t}\rangle) \notag\\
		&=
		\langle \bm{u}^r_{1,1},\frac{1}{2}\bm{e}_{h,1},\frac{1}{2}\bm{e}_{t,1}\rangle
		+\langle \bm{u}^{r}_{1,1},\frac{1}{2}\bm{e}_{h,2},\frac{1}{2}\bm{e}_{t,2}\rangle 
		+\langle \bm{u}^{r}_{2,1},\frac{1}{2}\bm{e}_{h,1},\frac{1}{2}\bm{e}_{t,2}\rangle \notag\\
		&\quad -\langle \bm{u}^{r}_{2,1},\frac{1}{2}\bm{e}_{h,2},\frac{1}{2}\bm{e}_{t,1}\rangle \notag\\
		&= 	\sum\nolimits_{i=1}^{2}
		\sum\nolimits_{j=1}^{1} 
		\sum\nolimits_{k=1}^{2}  
		\!\!\omega_{ijk} \cdot \left\langle 
		\bm{u}^{r}_{ij}, {{\bm{P}}^{r,\text{CE}}_{ijk}{[1,:]}}\bm{E}_h, {\bm{P}}^{r,\text{CE}}_{ijk}[2,:]\bm{E}_{t} 
		\right\rangle, \notag
	\end{align*}
where $\bm{r}=\bm{u}^r_{1,1}+\bm{u}^r_{2,1}\bm{{\rm i}}$, $\bm{h}=\bm{e}_{h,1}+\bm{e}_{h,2}\bm{{\rm i}}$, and $\bm{t}=\bm{e}_{t,1}+\bm{e}_{t,2}\bm{{\rm i}}$ are complex vectors and $\text{Re}(\cdot)$ is to take the real part. The pattern matrices are expressed as:
	\begin{equation}
		\bm{P}_{1,1,1}^{r,\text{CE}}\!\!=\!\!{
			\left[ \arraycolsep=2pt\def\arraystretch{1}
			\begin{array}{cc}
				\nicefrac{1}{2} & 0  \\
				\nicefrac{1}{2} & 0	
			\end{array} 
			\right ]},
		\bm{P}_{1,1,2}^{r,\text{CE}}\!\!=\!\!{
			\left[ \arraycolsep=2pt\def\arraystretch{1}
			\begin{array}{cc}
				0 & \nicefrac{1}{2} \\
				0 & \nicefrac{1}{2} 	
			\end{array} 
			\right ]},\notag
		 \\ \\
		\bm{P}_{2,1,1}^{r,\text{CE}}\!\!=\!\!{
			\left[ \arraycolsep=2pt\def\arraystretch{1}
			\begin{array}{cc}
				\nicefrac{1}{2} & 0 \\
				0 & \nicefrac{1}{2}	
			\end{array} 
			\right ]},
		\bm{P}_{2,1,2}^{r,\text{CE}}\!\!=\!\!{
			\left[ \arraycolsep=2pt\def\arraystretch{1}
			\begin{array}{cc}
				0 & \nicefrac{1}{2} \\
				\nicefrac{1}{2} & 0	
			\end{array} 
			\right ]}.
	\end{equation}

As for generalizing to QuatE, RAM sets the multiplicity of entity embeddings $m$ to 4, e.g., $\bm{E}_h=[\bm{e}_{h,1};\bm{e}_{h,2};\bm{e}_{h,3};\bm{e}_{h,4}]$. The multiplicity of role embeddings $m_{\gamma}$ is set to 2, where each role embedding owns four pattern matrices $n_{\text{Pmat}}=4$, e.g., $\gamma^r_1$ with $\bm{u}^r_{1,1}$ and $\bm{u}^r_{1,2}$, $\bm{u}^r_{1,1}$ with $\bm{P}^{r,\text{QE}}_{1,1,i},i\in\{1,\cdots,4\}$. 
Based on \citep{zhang2019quaternion}, the scoring function of QuatE can be rewritten as:

\begin{small}
	\begin{align*}
		\phi&(r,h,t)
		=\frac{1}{4}\bm{Q}_h\otimes \bm{W}_r\cdot \bm{Q}_t \notag\\ 
		=&
		\langle\bm{u}^{r}_{1,1},\frac{1}{2}\bm{e}_{h,1},\frac{1}{2}\bm{e}_{t,1}\rangle +
		\langle\bm{u}^{r}_{1,1},\frac{1}{2}\bm{e}_{h,2},\frac{1}{2}\bm{e}_{t,2}\rangle
		+
		\langle\bm{u}^{r}_{1,1},\frac{1}{2}\bm{e}_{h,3},\frac{1}{2}\bm{e}_{t,3}\rangle
		 \notag\\
		\quad &+\langle\bm{u}^{r}_{1,1},\frac{1}{2}\bm{e}_{h,4},\frac{1}{2}\bm{e}_{t,4}\rangle\notag \\
		&+
		\langle\bm{u}^{r}_{1,2},\frac{1}{2}\bm{e}_{h,1},\frac{1}{2}\bm{e}_{t,2}\rangle
		-
		\langle\bm{u}^{r}_{1,2},\frac{1}{2}\bm{e}_{h,2},\frac{1}{2}\bm{e}_{t,1}\rangle
		+
		\langle\bm{u}^{r}_{1,2},\frac{1}{2}\bm{e}_{h,3},\frac{1}{2}\bm{e}_{t,4}\rangle
		\notag \\
		&
		-\langle\bm{u}^{r}_{1,2},\frac{1}{2}\bm{e}_{h,4},\frac{1}{2}\bm{e}_{t,3}\rangle
		\notag\\
		&+
		\langle\bm{u}^{r}_{2,1},\frac{1}{2}\bm{e}_{h,1},\frac{1}{2}\bm{e}_{t,3}\rangle -
		\langle\bm{u}^{r}_{2,1},\frac{1}{2}\bm{e}_{h,2},\frac{1}{2}\bm{e}_{t,4}\rangle
		-
		\langle\bm{u}^{r}_{2,1},\frac{1}{2}\bm{e}_{h,3},\frac{1}{2}\bm{e}_{t,1}\rangle
		\notag\\
		&
		+
		\langle\bm{u}^{r}_{2,1},\frac{1}{2}\bm{e}_{h,4},\frac{1}{2}\bm{e}_{t,2}\rangle\notag\\
		&+
		\langle\bm{u}^{r}_{2,1},\frac{1}{2}\bm{e}_{h,1},\frac{1}{2}\bm{e}_{t,4}\rangle
		+
		\langle\bm{u}^{r}_{2,2},\frac{1}{2}\bm{e}_{h,2},\frac{1}{2}\bm{e}_{t,3}\rangle
		-
		\langle\bm{u}^{r}_{2,2},\frac{1}{2}\bm{e}_{h,3},\frac{1}{2}\bm{e}_{t,2}\rangle
		\notag \\
		&-
		\langle\bm{u}^{r}_{2,2},\frac{1}{2}\bm{e}_{h,4},\frac{1}{2}\bm{e}_{t,1}\rangle
		\notag\\
		&= 	\sum\nolimits_{i=1}^{2}
		\sum\nolimits_{j=1}^{2} 
		\sum\nolimits_{k=1}^{4}  
		\!\!\omega_{ijk} \cdot \left\langle 
		\bm{u}^{r}_{ij}, {{\bm{P}}^{r,\text{QE}}_{ijk}{[1,:]}}\bm{E}_h, {\bm{P}}^{r,\text{QE}}_{ijk}[2,:]\bm{E}_{t} 
		\right\rangle, \notag
	\end{align*}
\end{small}
where $\bm{W}_r=\bm{u}^{r}_{1,1}+\bm{u}^{r}_{1,2}\bm{{\rm i}}+\bm{u}^{r}_{2,1}\bm{{\rm j}}+\bm{u}^{r}_{2,2}\bm{{\rm k}},\ \bm{Q}_h=\bm{e}_{h,1}+\bm{e}_{h,2}\bm{{\rm i}}+\bm{e}_{h,3}\bm{{\rm j}}+\bm{e}_{h,4}\bm{{\rm k}},\ \bm{Q}_t=\bm{e}_{t,1}+\bm{e}_{t,2}\bm{{\rm i}}+\bm{e}_{t,3}\bm{{\rm j}}+\bm{e}_{t,4}\bm{{\rm k}}$ are quaternion vectors, $\otimes$ and $\cdot$ are Hamilton product and inner product respectively. 
\begin{figure*}[hbt!]
	\centering
	\subfigure[FB15k/children]{
		\includegraphics[width=0.23\textwidth]{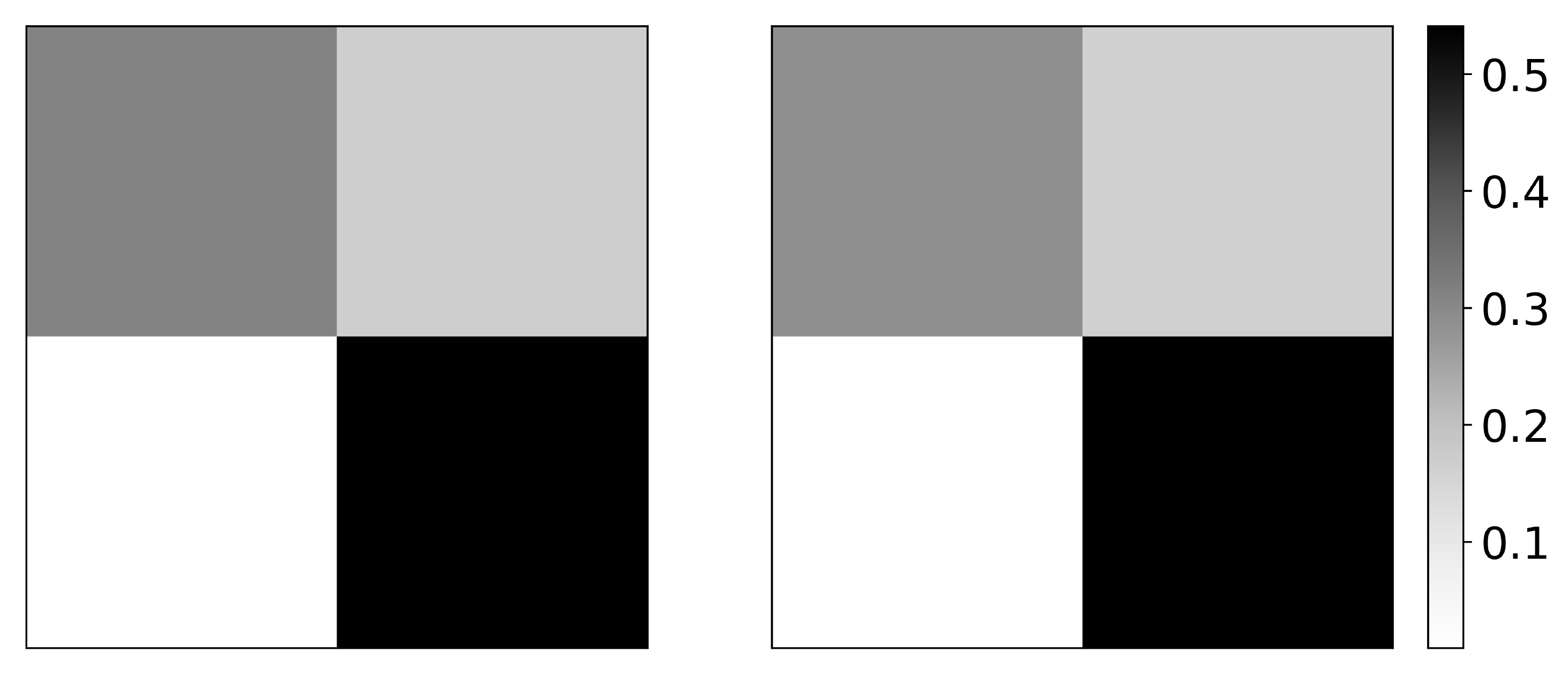}}
	\subfigure[FB15k/parent]{
		\includegraphics[width=0.23\textwidth]{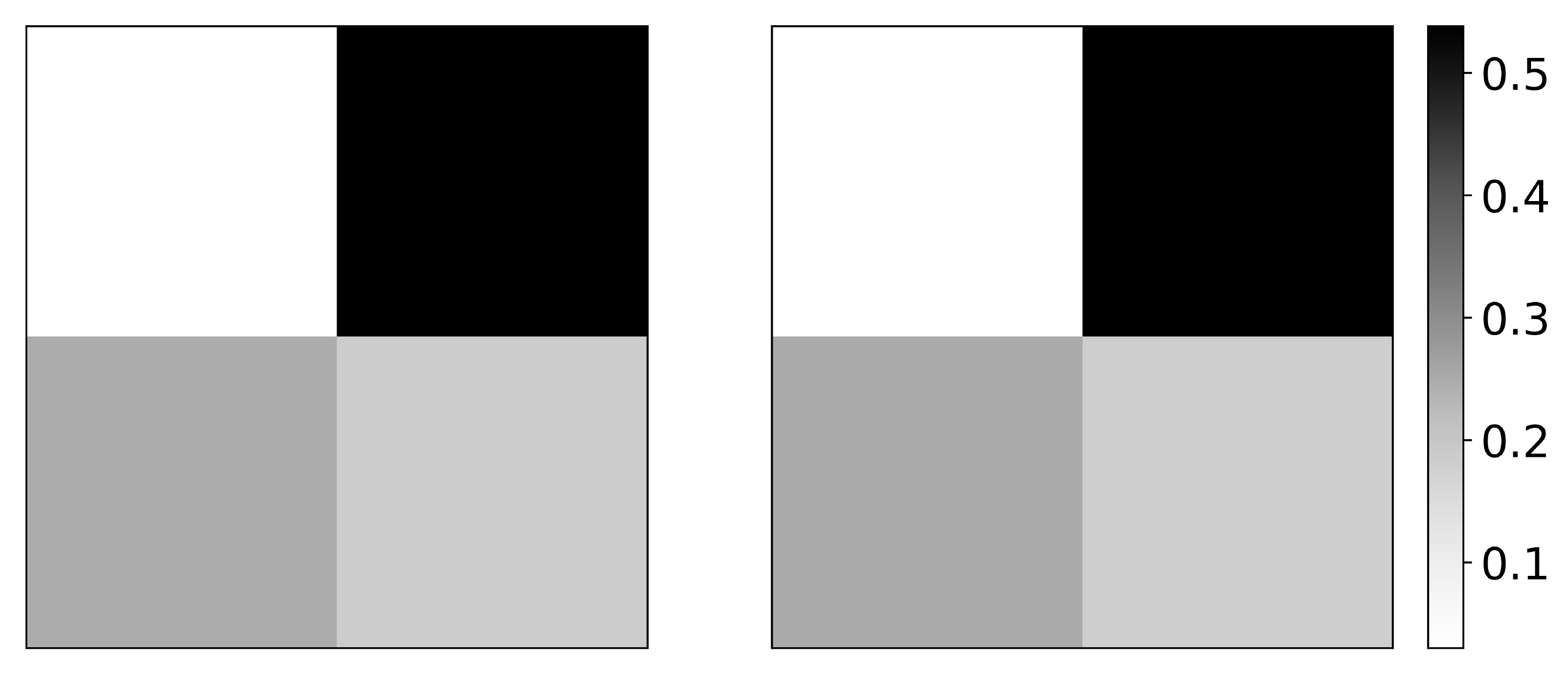}}
	\subfigure[WikiPeople/father]{
		\includegraphics[width=0.23\textwidth]{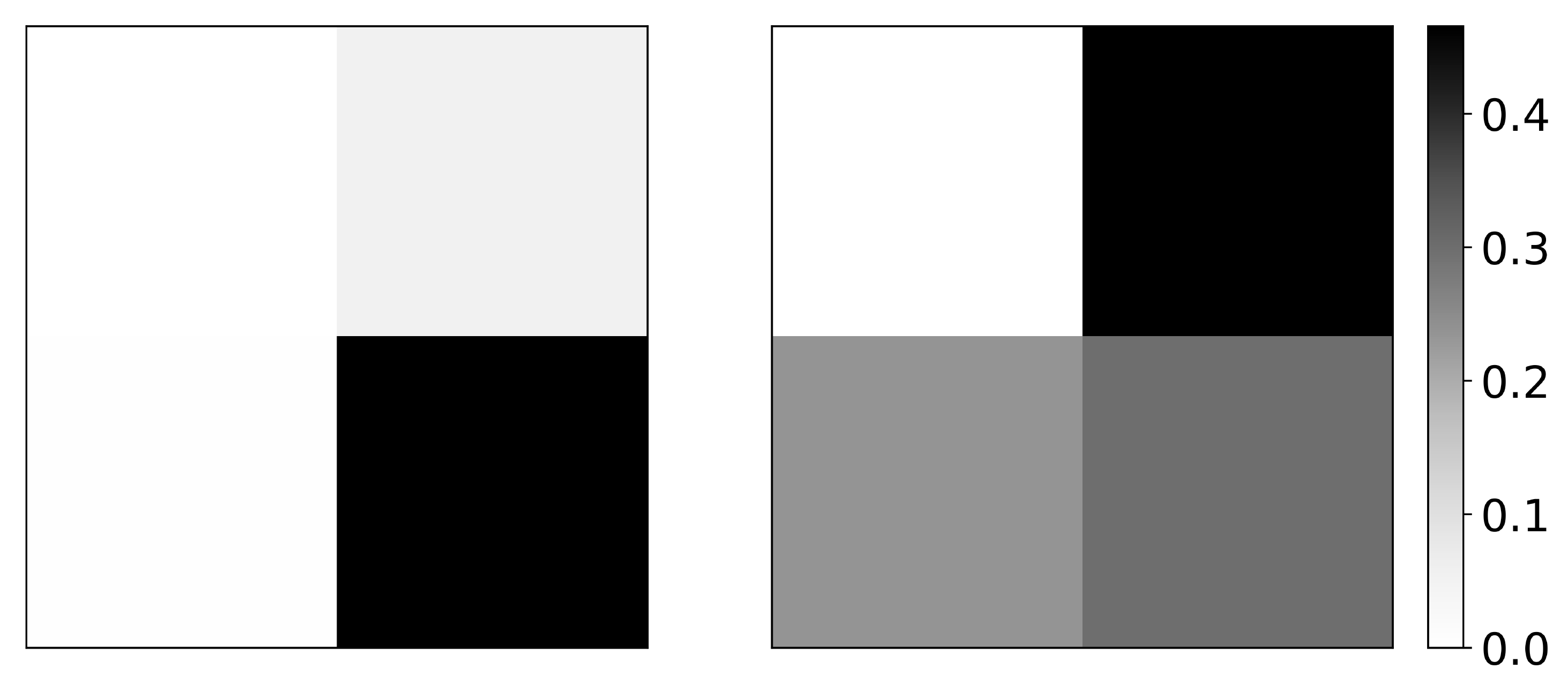}}
	\subfigure[WikiPeople/mother]{
		\includegraphics[width=0.23\textwidth]{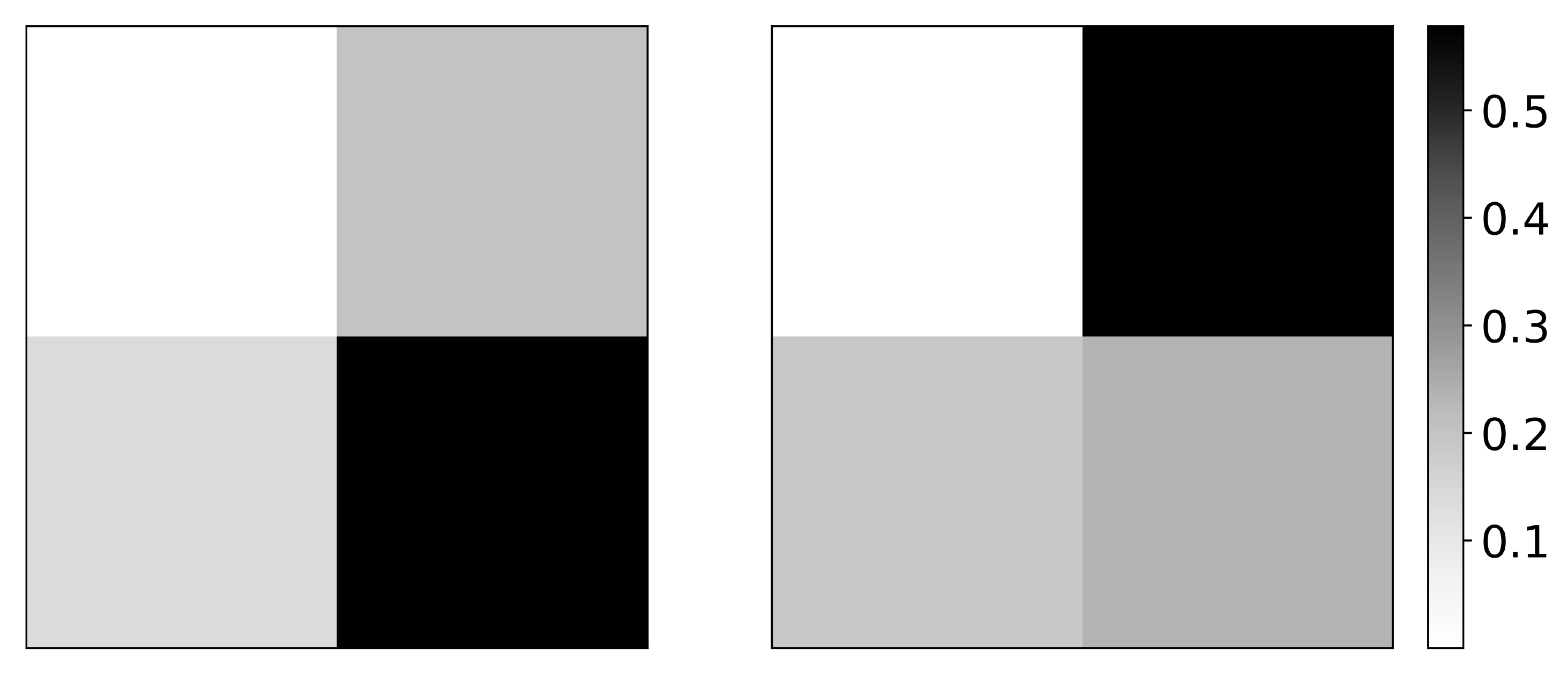}}
	\vspace{-10px}
	\caption{Role-aware pattern matrix visualization of binary relations in FB15k and WikiPeople, in respect of inverse relations and similar relations.}\label{fig:vis-app}
\end{figure*}

Due to space limitation, 
we only illustrate the pattern matrices for the embedding vectors of the first role $\text{rol}^r_1$:
	\begin{align*}
		&	\bm{P}_{1,1,1}^{r,\text{QE}}={
			\left[ \arraycolsep=1.5pt\def\arraystretch{1}
			\begin{array}{cccc}
				\nicefrac{1}{2} & 0 & 0 & 0 \\
				\nicefrac{1}{2} & 0	& 0 & 0
			\end{array} 
			\right]},
		&	\bm{P}_{1,1,2}^{r,\text{QE}}={
			\left[ \arraycolsep=1.5pt\def\arraystretch{1}
			\begin{array}{cccc}
				0 & \nicefrac{1}{2} & 0 & 0 \\
				0 & \nicefrac{1}{2}	& 0 & 0
			\end{array} 
			\right]}, \notag 
		\\
		&	\bm{P}_{1,1,3}^{r,\text{QE}}={
			\left[ \arraycolsep=1.5pt\def\arraystretch{1}
			\begin{array}{cccc}
				0 & 0 & \nicefrac{1}{2} & 0 \\
				0 & 0 & \nicefrac{1}{2} & 0
			\end{array} 
			\right]},
		&	\bm{P}_{1,1,4}^{r,\text{QE}}={
			\left[ \arraycolsep=1.5pt\def\arraystretch{1}
			\begin{array}{cccc}
				0 & 0 & 0 & \nicefrac{1}{2} \\
				0 & 0	& 0 & \nicefrac{1}{2}
			\end{array} 
			\right]},\notag
		\\
		&		\bm{P}_{1,2,1}^{r,\text{QE}}={
			\left[ \arraycolsep=1.5pt\def\arraystretch{1}
			\begin{array}{cccc}
				\nicefrac{1}{2} & 0 & 0 & 0 \\
				0 & \nicefrac{1}{2}	& 0 & 0
			\end{array} 
			\right]},
		&	\bm{P}_{1,2,2}^{r,\text{QE}}={
			\left[ \arraycolsep=1.5pt\def\arraystretch{1}
			\begin{array}{cccc}
				0& \nicefrac{1}{2} & 0 & 0 \\
				\nicefrac{1}{2} & 0	& 0 & 0
			\end{array} 
			\right]}, \notag\\
		&	\bm{P}_{1,2,3}^{r,\text{QE}}={
			\left[ \arraycolsep=1.5pt\def\arraystretch{1}
			\begin{array}{cccc}
				0 & 0 & \nicefrac{1}{2} & 0 \\
				0 & 0	& 0 & \nicefrac{1}{2}
			\end{array} 
			\right]},\notag
		&	\bm{P}_{1,2,4}^{r,\text{QE}}={
			\left[ \arraycolsep=1.5pt\def\arraystretch{1}
			\begin{array}{cccc}
				0 & 0 & 0 & \nicefrac{1}{2} \\
				0 & 0	& \nicefrac{1}{2} & 0
			\end{array} 
			\right]}.\notag
	\notag	
	\end{align*}

\section{Experimental Details}\label{app:exp}

\subsection{Role-Aware Pattern Matrix Visualization}\label{app:vis}
Here we provide some typical visualization results on binary relational dataset FB15k and n-ary relational dataset WikiPeople. 

In Figure~\ref{fig:vis-app}(a) and (b), we visualize the role-aware pattern matrices of an inverse relation pair, i.e., \texttt{children} and \texttt{parent}. 
It can be observed that exchanging rows of \texttt{children}'s pattern matrices roughly leads to \texttt{parent}'s pattern matrices, 
which are in accord with the statement in main contents. 
Moreover, the pattern matrices of similar relations of \texttt{father} and \texttt{mother} in WikiPeople are shown in Figure~\ref{fig:vis-app}(c) and (d). 
Since these two relations are semantically related, 
and always involve similar entities in n-ary relational KBs, 
their role-aware pattern matrices are close to each other.
We believe such visualization results provide insight to explainable n-ary relational KB modeling especially embedding techniques.

\subsection{Influence of Binary Relational Data Ratio}
\textbf{Analysis on WikiPeople.} According to Table~\ref{tab:mbkbin} in main contents, RAM/b achieves better performance than RAM on binary relational data of WikiPeople. That is mainly because binary relational data accounts for over 88\% of training set in WikiPeople, which is enough for training a good RAM/b model. On the other hand, due to the unbalanced distribution of binary and beyond-binary relational data, n-ary relational data might introduce more noise than gains for training RAM for binary relational KBC. 

\begin{figure}[ht]
	\centering
	\includegraphics[width=0.38\textwidth]{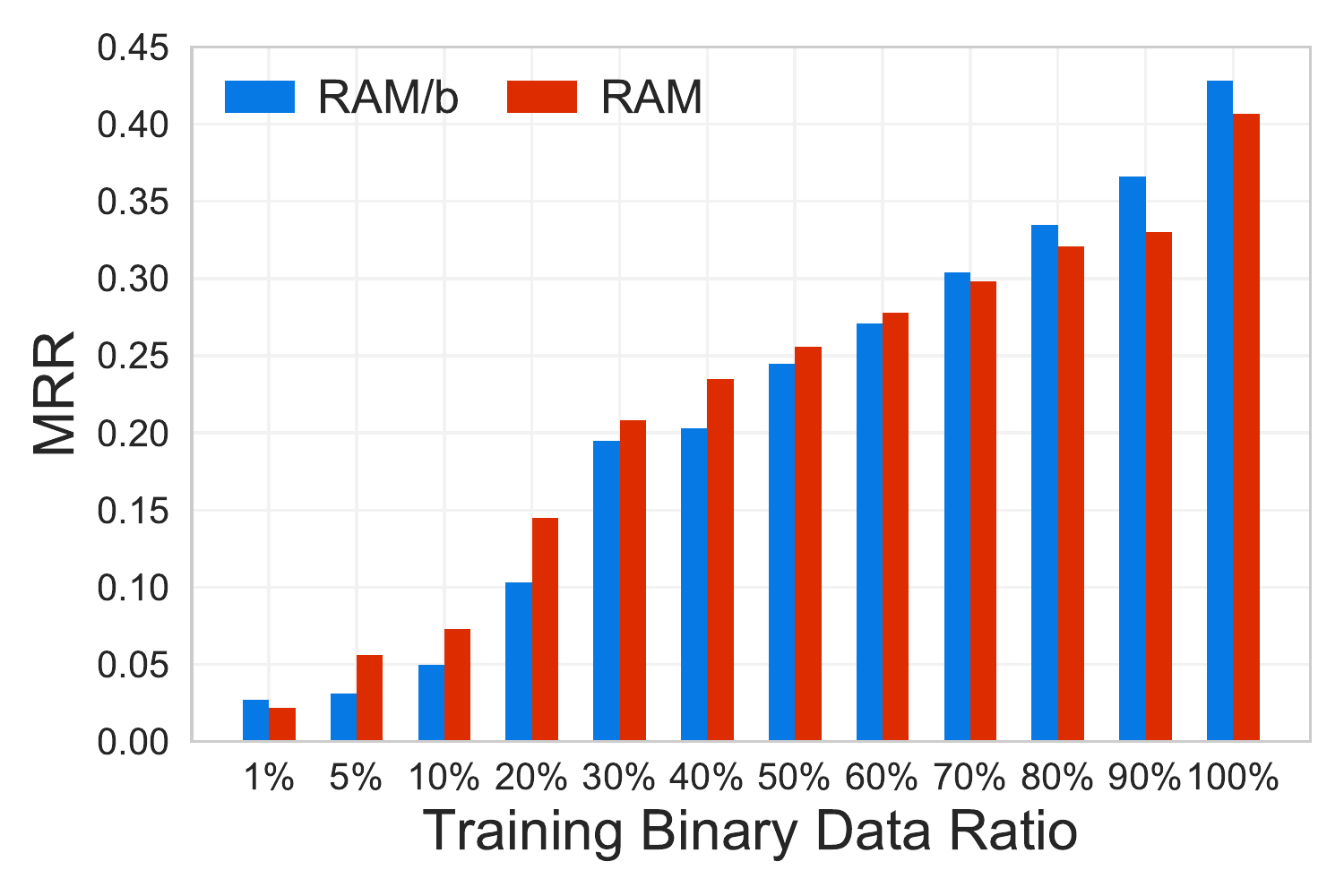}
	\vspace{-10px}
	\caption{Binary relational KBC results on WikiPeople under different ratios of training binary data. $x$-axis identifies the ratio of training binary data compared with the original WikiPeople dataset. RAM is trained with n-ary relational facts while RAM/b is trained with binary relational facts only. Both models are tested with binary relational facts in original testing dataset.}\label{fig:bdratio}
\end{figure}

To analyze the phenomenon, we randomly drop a part of binary relational data in training set of WikiPeople, and train RAM/b and RAM on the processed dataset. The learned models are evaluated on original testing binary relational data of WikiPeople. Figure~\ref{fig:bdratio} shows the performance of both models trained on different ratios of binary relational data. It can be observed that, RAM achieves better performance when the training binary data is few (under 60\%), which shows the positive gains brought by information sharing across roles. As for enough training binary relational data (over 70\%), RAM is more sensitive to noise from n-ary relational data, and thus leads to worse performance than RAM/b. Such results validate the analysis above, and demonstrate the importance of n-ary relational data for modeling the sparse binary relational KBs.

\bibliographystyle{ACM-Reference-Format}
\balance
\bibliography{ref}
%

\end{document}